\documentclass[10pt,journal,twoside,onecolumn]{IEEEtran}

\usepackage{psfrag,epsfig,graphics}
\usepackage{amsmath,amssymb,multirow}
\usepackage{setspace}

\usepackage{graphicx}
\usepackage{caption}
\usepackage{subcaption}

\usepackage[noadjust]{cite}
\usepackage{multirow}
\usepackage[lined,linesnumbered,ruled]{algorithm2e}

\usepackage{color}  

\def\cblue{\textcolor{blue}}

\newcommand{\cb}[1]{{\boldsymbol{#1}}}
\newcommand{\cp}[1]{\ifmmode {\mathcal{#1}}\else ${\mathcal{#1}}$\fi}
\newcommand{\balpha}{\boldsymbol{\alpha}}

\newcommand{\bkappa}{\boldsymbol{\kappa}}

\newcommand{\bzero}{\boldsymbol{0}}

\newcommand{\bI}{\boldsymbol{I}}
\newcommand{\bK}{\boldsymbol{K}}

\newcommand{\bR}{\boldsymbol{R}}

\newcommand{\bm}{\boldsymbol{m}}
\newcommand{\bn}{\boldsymbol{n}}

\newcommand{\bp}{\boldsymbol{p}}
\newcommand{\be}{\boldsymbol{e}}

\newcommand{\br}{\boldsymbol{r}}

\newcommand{\bnu}{\boldsymbol{\nu}}

\newcommand{\mr}[1]{{\bm_{\lambda_{#1}}}}

\newcommand{\MM}{{\boldsymbol{M}}}

\newcommand{\KK}{\boldsymbol{K}}

\begin{document}

\title{\LARGE{Nonparametric Detection of Nonlinearly Mixed Pixels \\ and Endmember Estimation in Hyperspectral Images}}
\author{Tales Imbiriba$^{(1)}$, \IEEEmembership{Student Member, IEEE}, Jos{\'e} Carlos Moreira Bermudez$^{(1)}$, \IEEEmembership{Senior Member, IEEE} \\ C{\'e}dric Richard$^{(2)}$, \IEEEmembership{Senior Member, IEEE}, Jean-Yves Tourneret $^{(3)}$, \IEEEmembership{Senior Member, IEEE}
\thanks{The work of J.-C. M. Bermudez was partly supported by CNPq grants 305377/2009-4, 400566/2013-3 and 141094/2012-5. The work of C. Richard and J.-Y. Tourneret was partly supported by ANR grants ANR-12- BS03-003 (Hypanema) and ANR-11-LABX-0040-CIMI.
This work appeared in part in the Proceedings of the IEEE International Conference on Acoustics, Speech, and Signal Processing (ICASSP), Florence, Italy, May 2014~\cite{imbiriba2014detection}.}
 \\ \vspace{0.5cm}
\small{$^{(1)}$ Federal University of Santa Catarina \\
Department of Electrical Engineering, 88040-900, Florian{\'o}polis, SC - Brazil \\
tel.: +55.48.3721.7719 \hspace{0.5cm} \hspace{0.5cm} fax.:
+55.48.3721.9280 \\ talesim@gmail.com \hspace{0.5cm} j.bermudez@ieee.org}\vspace{0.3cm}\\
\small{\linespread{0.2} $^{(2)}$ Laboratoire Lagrange \\
Universit{\'e} de Nice Sophia-Antipolis, France \\
phone: (33) 492 076 394 \hspace{0.5cm} \hspace{0.5cm} fax:
(33) 492 076 321 \\
cedric.richard@unice.fr\vspace{0.3cm} \\
$^{(3)}$ Universit{\'e} de Toulouse, IRIT/INP-ENSEEIHT/TéSA, France \\
phone: (33) 534 322 224 \\
jean-yves.tourneret@enseeiht.fr}
}
\maketitle
\begin{abstract}
Mixing phenomena in hyperspectral images depend on a variety of factors such as the resolution of observation devices, the properties of materials, and how these materials interact with incident light in the scene. Different parametric and nonparametric models have been considered to address hyperspectral unmixing problems. The simplest one is the linear mixing model. Nevertheless, it has been recognized that mixing phenomena can also be nonlinear. The corresponding nonlinear analysis techniques are necessarily more challenging and complex than those employed for linear unmixing. Within this context, it makes sense to detect the nonlinearly mixed pixels in an image prior to its analysis, and then employ the simplest possible unmixing technique to analyze each pixel. In this paper, we propose a technique for detecting nonlinearly mixed pixels. The detection approach is based on the comparison of the reconstruction errors using both a Gaussian process regression model and a linear regression model. The two errors are combined into a detection statistics for which a probability density function can be reasonably approximated. We also propose an iterative endmember extraction algorithm to be employed in combination with the detection algorithm. The proposed Detect-then-Unmix strategy, which consists of extracting endmembers, detecting nonlinearly mixed pixels and unmixing, is tested with synthetic and real images.
\end{abstract}

\maketitle

\newpage

\section{Introduction}

Emerged in the 1960s with multispectral scanners, modern hyperspectral sensors produce two-dimensional hyperspectral images over a few tens to thousands of contiguous spectral bands~\cite{Bioucas-Dias-2013-ID307}. Their high spectral resolution allows a comprehensive and quantitive analysis of materials in remotely observed data. This area has received considerable attention in the last decade. Due to historic downlink and computer processing limitations~\cite{Landgrebe1997}, hyperspectral images often trade spatial for spectral resolution~\cite{Landgrebe:2002p5665}. The observed reflectances then result from spectral mixtures of several pure material signatures. As a consequence, spectral unmixing has become an important issue for hyperspectral data processing~\cite{Keshava:2002p5667}.

In a supervised setting, the spectral signatures of pure materials are available as vectors of reflectances of these materials at each wavelength. Such vectors are typically called endmembers due to their geometrical interpretation in the linear mixing case. Mixing phenomena depend on a variety of factors such as the resolution of observation devices, the properties of materials, and how these materials interact with incident light in the scene~\cite{Dobigeon-2014-ID322}. Therefore, different parametric and nonparametric models have been considered to address hyperspectral unmixing problems. The simplest one is the linear mixing model, which assumes linear mixing of the endmembers contributions~\cite{Keshava:2002p5667}. It has been recognized that mixing phenomena can also be nonlinear~\cite{Keshava:2002p5667, Dobigeon-2014-ID322}. The corresponding analysis techniques are necessarily more challenging and complex than those employed for linear unmixing. Nevertheless, nonlinear analysis of hyperspectral images has been widely explored in the past few years. See, for instance,~\cite{Dobigeon-2014-ID322,Bioucas-Dias-2013-ID307, Ray:1996tp, Nascimento2009, Somers:2009p6577, guilfoyle2001,fan2009,altmann2011,chen2013nonlinear,chen2012nonlinear,chen2013estimating,chen2014nonlinear2}. Nonlinear unmixing algorithms can lead to a better understanding of the individual spectral contributions, despite the increased complexity. Hence, it makes sense to detect the nonlinearly mixed pixels in an image prior to its analysis, and then employ the simplest possible unmixing technique to analyze each pixel. To this end, it is desirable to devise analysis techniques that combine endmember extraction, detection of nonlinearly mixed pixels and unmixing.

The problems of extracting endmembers, detecting nonlinearly mixed pixels and unmixing are interlaced, and addressing them jointly is not a trivial task.  For instance, most nonlinear unmixing techniques assume the endmembers to be known or to be estimated by an endmember extraction algorithm~\cite{chen2013nonlinear,chen2012nonlinear,chen2013estimating,chen2014nonlinear2,Altmann:2011vb,halimi2011,Broadwater2007,Broadwater2009,Altmann-2014-ID337}. However, most endmember extraction algorithms rely on the convex geometry associated with the linear mixing model~\cite{Boardman1993,Nascimento2005,chang2006new,Chan-2009-ID334,Chan:2011:SVMAX}, which obviously does not apply to nonlinearly mixed pixels. Endmember extraction techniques designed for situations where a significant part of the image is composed of nonlinear mixtures are rarely addressed in the literature. In fact, most of the techniques considering nonlinearly mixed pixels are part of a complete unsupervised unmixing strategy~\cite{Heylen2011nonlinear, Altmann-2013-ID311}. Detecting nonlinearly mixed pixels in an hyperspectral image is also a complex task. Physically motivated models \cite{Borel:1994tp,Ray:1996tp} usually tend to be too complex for application in practical detection strategies. One possible approach is to consider a simplified parametric model for the nonlinearity.  The parameters of this nonlinear model are then estimated from the image, and hypothesis tests are derived based on these estimates. For instance, a single-parameter polynomial post-nonlinear model is assumed in~\cite{Altmann-2013-ID308}. The main question regarding parametric modeling of nonlinear mixing mechanisms is whether the chosen model can capture the actual nonlinear effects present in a scene. When nothing or little is known about the nonlinear mixing mechanism, a direct strategy is to exploit the property of linear mixing models to confine the noiseless data to a simplex. The hypothesis test proposed in~\cite{Altmann-2013-ID306} is based on the distance between the observed pixel and this simplex. Though this test is robust to nonlinear mixing mechanisms, it conveys too little information about the nonlinearity as a tradeoff to guarantee simplicity. An alternative strategy is to use nonparametric techniques to extract information about the nonlinearity directly from the observations. A nonparametric unmixing technique based on kernel expansions is presented in~\cite{chen2013nonlinear}, but this work does not address nonlinearity detection. A nonlinear mixing model for joint unmixing and nonlinearity detection is proposed in~\cite{Altmann-2014-ID335}. It assumes that the observed reflectances result from linear spectral mixtures corrupted by a residual nonlinear component. This model is rather similar to the model initially introduced in~\cite{chen2013nonlinear}, but the estimation method relies on a computationally intensive Bayesian procedure.

All the detection methods discussed above assume known endmember spectral signatures. In most cases, the endmembers are assumed to have been estimated from the data. However, most endmember extraction algorithms exploit the convex geometry of the linear mixing model and assume the presence of pure endmember pixels in the image. They usually exploit one of the following properties: 1) the endmembers are the extreme points when projecting the data onto any subspace~\cite{Boardman1993,Nascimento2005,chang2006new,Chan:2011:SVMAX,gillis2014fast,honeine2012geometric}; 2) the volume of a simplex spanned by any subset of points in the image is maximum when these points are the endmembers~\cite{Winter1999,neville1999automatic}. Other strategies deal with nonnegative matrix factorization~\cite{ammanouil2014blind,ammanouil2014glup,miao2007endmember}. One method of particular interest to this work is the MVES (minimum volume enclosing simplex) algorithm~\cite{Chan-2009-ID334} as it does not assume the presence of pure pixels, though it still exploits the geometry of linear mixtures. It solves a constrained least-squares optimization problem with a simplex volume regularizer.  Practically, MVES finds the smallest simplex circumscribing the hyperspectral data. The vertices of this simplex are defined as the endmembers.  

In this paper, we propose a technique that combines endmember extraction and detection of nonlinearly mixed pixels in hyperspectral images. The detection approach is based on the comparison of the reconstruction errors using both a Gaussian process (GP) and a linear regression model. The two errors are combined into a detection statistics for which a probability density function can be reasonably approximated. We also propose an MVES-based iterative endmember extraction algorithm to be employed in combination with the detection algorithm to jointly detect nonlinearly mixed pixels and extract the image endmembers. The proposed method is tested with synthetic and real images. This work is organized as follows. Section~\ref{sec:MM} reviews the linear mixing model and some nonlinear mixing models, and discusses different ways of modeling nonlinear interaction between light and endmembers. Section~\ref{sec:NLD} discusses GP regression applied in the context of hyperspectral data unmixing. Section~\ref{sec:Detection} presents the application of GP to the detection of nonlinearly mixed pixels. Section~\ref{sec:EMExtraction} introduces a two-step iterative procedure to estimate the endmember matrix. This method combines the MVES algorithm and the nonlinear mixture detector proposed in Section~\ref{sec:Detection}. Simulations with synthetic and real data are presented in Section~\ref{sec:Sim}. Conclusions are finally presented in Section~\ref{sec:conc}.

\section{Mixture models}
\label{sec:MM}

Each observed pixel can be written as a function of the endmembers plus an additive term associated with the measurement noise and the modeling error. Consider the model:
\begin{equation}
       \label{eq:model}
        \br = {\cb\psi}(\MM) + \cb{n}
\end{equation}
where $\br =[r_1,\ldots, r_L]^\top$ is a vector of reflectances observed in $L$ spectral bands, $\MM = [\bm_1,\ldots,\bm_R]$ is the $L\times R$ endmember matrix, whose $i$-th column $\bm_i$ is an endmember, $\bn\sim\cp{N}(\bzero,\sigma_n^2\bI)$ is a white Gaussian noise  (WGN) vector, and $\cb{\psi}$ is an unknown mixing function. Several models of the form \eqref{eq:model} have been proposed in the literature, depending on the linearity or nonlinearity of $\cb{\psi}$, type of mixture, and other properties~\cite{Dobigeon-2014-ID322}.

\subsection{The linear mixing model}

The linear mixing model assumes that each light ray interacts only with one material, disregarding multiple interactions between light and multiple materials~\cite{Keshava:2002p5667}. The classical linear model assumes that $\cb{\psi}$ is a convex combination of the endmembers. In this situation, the vector $\br$ can then be written as  
\begin{equation}
\begin{split}
	&\br = \MM \balpha + \bn\\
	&\text{subject to }\,\cb{1}^\top\balpha = 1 \text{ and } \balpha \succeq \cb{0}
 \label{eq:linForm}
\end{split}
\end{equation}
 where $\balpha = [\alpha_1,\ldots,\alpha_{R}]^\top$ is the vector of abundances of each endmember in $\MM$, $R$ is the number of endmembers, and $\succeq$ denotes the entrywise $\geq$ operator. Therefore, the entries of $\balpha$ cannot be negative and should sum to one. The observation $r_\ell$ in the $\ell$-th wavelength of~\eqref{eq:linForm} can be written as 
 \begin{equation}
  r_\ell = \bm_{\lambda_\ell}^\top\balpha + n_\ell
  \label{eq:LMM_band}
 \end{equation}
where $\bm_{\lambda_\ell}$ denotes the $\ell$-th row of $\MM$ as a column vector.
In the noiseless case, namely, $n_\ell = 0$, the sum-to-one and positivity constraints over the abundances in~\eqref{eq:linForm} confine the data to a simplex. The vertices of this simplex are the endmembers, which justifies the terminology.

Several parametric models have been proposed in the literature to describe nonlinear mixing mechanisms of endmembers in hyperspectral images. See~\cite{Dobigeon-2014-ID322} and references therein. We shall now review two popular models that will be used later to generate synthetic data for evaluation purposes.

\subsection{Nonlinear mixing models}

The generalized bilinear model (GBM)~\cite{halimi2011} is given by
\begin{equation}
	\label{eq:GBM_orig}
	\begin{split}
		&\br = \MM\balpha + \sum_{i=1}^{R-1}\sum_{j=i+1}^{R}\gamma_{ij}\,\alpha_i\alpha_j\,\bm_i\odot\bm_j + \bn \\
		&\text{subject to }\cb{1}^\top\balpha = 1 \text{ and } \balpha \succeq 0
	\end{split}
\end{equation}
where the parameters $\gamma_{ij}\in[0,1]$ govern the amount of nonlinear contribution, and $\odot$ denotes the Hadamard product.  In the noiseless case, data following the model~\eqref{eq:GBM_orig} lie in a nonlinearly distorted simplex in $\mathbb{R}^R$ whose vertices are the endmembers as in the linear case. For simplicity, here we consider a simplified version of this model in which the nonlinear contribution is controlled by a single parameter $\gamma$ such that $\gamma = \gamma_{ij}$ for all $(i,j)$.

The post nonlinear mixing model (PNMM)~\cite{Jutten2003} is given by
\begin{equation}
 \br = \cb{g}(\MM\balpha) + \bn
 \label{eq:PostLinearModel}
\end{equation}
where $\cb{g}$ is a nonlinear function applied to the linear mixing model. The PNMM can represent a wide range of nonlinear mixing models, depending on the definition of $\cb{g}$. For instance, the PNMM considered in~\cite{chen2013nonlinear} is given by
\begin{equation}
	\label{eq:pnmm_chen}
	\br = (\MM\balpha)^\xi + \bn
\end{equation}
where $(\cb{v})^{\xi}$ denotes the exponentiation applied to each entry of the vector $\cb{v}$. For $\xi=2$, \eqref{eq:pnmm_chen} becomes a bilinear model closely related to the GBM but without a linear term.  The PNMM was explored in other works considering different forms for $\cb{g}$ applied to hyperspectral data unmixing~\cite{Altmann-2013-ID308,altmann2011:icassp}. 

The GBM~\eqref{eq:GBM_orig} and the PNMM~\eqref{eq:PostLinearModel} nonlinear mixing models mainly represent the scattering phenomenon where the light first interacts with an endmember, and then with a second one, before being captured by the hyperspectral sensor. Other models account for other kinds of interaction between light and endmembers, or consider other types of nonlinear effects. In the case of the intimate mixture model~\cite{HapkeBook1993} for instance, the endmembers are considered to be mixed at the molecular level. Other nonlinear models can be considered depending on the characteristics of the scene~\cite{fan2009, Nascimento2009, Jutten2003, halimi2011,HapkeBook1993,Borel:1994tp,Somers:2009p6577,  Ray:1996tp, Broadwater2009}. More importantly, these informations are usually missing. Hence, it makes sense to develop nonparametric models that do not make strong assumptions about the type of nonlinearity involved in the mixture.

\section{Nonlinearity detection with Gaussian process regression models}
\label{sec:NLD}

To detect nonlinearly mixed pixels in an hyperspectral image, assuming $\cb{\psi}$ in \eqref{eq:model} is unknown, we propose to compare the reconstruction errors resulting from estimating $\cb{\psi}$ with nonlinear and linear regression methods.  Gaussian process (GP) regression methods consist of defining stochastic models for functions and performing inference in functional spaces~\cite{Rasmussen-2006-ID292}. The representation is rigorous but, at the same time, lets the data speak for themselves. This characteristic is desirable when little is known about the functions to be estimated. Using some knowledge obtained from the observations about the endmember matrix, we propose a supervised learning strategy to make inference on $\cb{\psi}$.

This section describes the application of GP nonlinear regression to the problem at hand.  Consider the training set $\{\MM,\br\}$ with inputs $\MM = [\mr{1},\ldots, \mr{L}]^\top$, and outputs or observations $\br = [r_1,\ldots,r_L]^{\top}$.  By analogy with the linear mixing model~\eqref{eq:LMM_band}, we write the $\ell$-th row of~\eqref{eq:model} as 
\begin{equation}
	r_\ell = \psi(\mr{\ell}) + n_\ell,
\end{equation}
with $r_\ell$ the $\ell$-th entry of the observation $\br$, $\psi$ a real-valued function in a (reproducing kernel) Hilbert space $\cp{H}$, and $n_\ell$ an additive WGN in the $\ell$-th band. A Gaussian process is a collection of random variables, any finite number of which has a joint Gaussian distribution~\cite{Rasmussen-2006-ID292}. We define a Gaussian prior distribution for $\psi$ with mean and covariance functions given by
\begin{equation}
	\begin{split}
		\mathbb{E}\{\psi(\mr{\ell})\} &= 0 \\
		\mathbb{E}\{\psi(\mr{\ell})\psi(\mr{\ell'})\} &= \kappa(\mr{\ell},\mr{\ell'})
	\end{split}
\end{equation}
where  $\kappa$ is a positive definite kernel. For notational simplicity, it is common but not necessary to consider GPs with a zero mean function. This assumption is not overly restricting as the mean of the posterior distribution is not confined to be zero (as shown by~\eqref{eq:predictiveGP}). The prior on the noisy observation $\br$ becomes:
\begin{equation}
	\label{eq:gp1}
	\br\sim\mathcal{N}(\cb{0},\bK+\sigma_{n}^2 \bI_L), 
\end{equation}
with $\bK$ the Gram matrix whose entries $\bK_{ij} = \kappa(\mr{i},\mr{j})$ are given by the kernel covariance function evaluated at $\mr{i}$ and $\mr{j}$, $\sigma_{n}^2$ the noise power, and $\bI_L$ the $L\times L$ identity matrix.

To obtain the predictive distribution for $\psi_*\triangleq\psi(\mr{*})$ at any test point $\mr{*}$, we can write the joint distribution of the observation $\br$ and $\psi(\mr{*})$ as~\cite{Rasmussen-2006-ID292}
\begin{equation}
	\label{eq:GPDist}
	\left[
	\begin{array}{c}
		\br\\
		\psi_*
	\end{array}
	\right]\! \sim \mathcal{N}\left(\cb{0}, 
  	\left[
  	\begin{array}{c c}
      	\bK + \sigma_{n}^2\bI_L  & \bkappa_* \\
      	\bkappa_*^{\top} & \kappa_{**}
  	\end{array}
  	\right]
  	\right) 
\end{equation}
with $\bkappa_{*} = [\kappa(\mr{*},\mr{1}), \ldots, \kappa(\mr{*},\mr{L})]^{\top}$ and $\kappa_{**} = \kappa(\mr{*},\mr{*})$. The predictive distribution of $\psi_*$, or posterior of $\psi_*$, is then obtained by conditioning~\eqref{eq:GPDist} on the observation as follows:
\begin{equation}
	\label{eq:predictiveGP}
	\psi_{*}|\br,\MM,\mr{*} \sim \mathcal{N}\left(\bkappa^{\top}_*\left[\bK +\sigma_{n}^2\bI_L\right]^{-1}\br,
	\kappa_{**} -\bkappa^{\top}_*\left[\bK+\sigma_{n}^2\bI_L\right]^{-1}\bkappa_*\right).
\end{equation}
The extension to a multivariate predictive distribution with test data $\MM_* = [\mr{* 1}, \ldots, \mr{* L}]^\top$ yields:
\begin{equation}
        \label{eq:predictiveGPVEC}
	\cb{\psi}_{*}|\br,\MM,\MM_* \sim \mathcal{N}\left(\bK^{\top}_*\left[\bK +\sigma_{n}^2\bI_L\right]^{-1}\br,
        \bK_{**} -\bK^{\top}_*\left[\bK+\sigma_{n}^2\bI_L\right]^{-1}\bK_*\right)
\end{equation}
where $[\bK_{*}]_{ij} = \kappa(\mr{\star i},\mr{j})$ and  $[\bK_{**}]_{ij} = \kappa(\mr{\star i},\mr{\star j})$. Finally, we arrive at the minimum mean square error (MMSE) estimator for GP regression:
\begin{equation}
	\label{eq:GPREst}
	\begin{split}
	\widehat{\cb{\psi}}_*&=\mathbb{E}\{\cb{\psi}_*|\br,\MM,\MM_*\} \\ &= \bK^{\top}_*\left[\bK +\sigma_{n}^2\bI_L\right]^{-1}\br.
	\end{split}
\end{equation}

In order to turn GP into a practical tool for processing hyperspectral data, it is essential to derive a method for estimating free parameters such as the noise variance $\sigma_n^2$ and possible kernel parameters defining the unknown parameter vector $\cb{\theta}$. We proceed as in~\cite{Rasmussen-2006-ID292} by maximizing the marginal  likelihood $p(\br| \MM,\sigma_n^2,\cb{\theta})$ with respect to $(\sigma_n^2,\cb{\theta})$, which leads to
\begin{equation}
	\label{eq:hyperparameter}
	(\hat{\sigma}_n^2,\hat{\cb{\theta}})=\arg\max_{\sigma_n^2,\cb{\theta}}\,\, \left(
			-\frac{1}{2}\br^{\top} \left[\KK + \sigma_{\!n}^2\bI_L\right]^{-1}\br - \frac{1}{2} \log |\KK + \sigma_{n}^2\bI_L|\right).
\end{equation}
This problem has to be addressed with numerical optimization methods. There is no guarantee that the cost function does not suffer from multiple local optima. However, our practical experience with hyperspectral data indicates that local optima are not a critical problem in this context. 

We conclude this section by introducing some kernel functions. Common examples include the linear kernel defined as:
\begin{equation}
	\label{eq:linear-kernel}
	\kappa(\mr{i},\mr{j}) = \cb{m}_{\lambda_i}^\top\cb{m}_{\lambda_j}^{\phantom{\top}}
\end{equation}
and radial basis function kernels, which depend on $\|\mr{i} -\mr{j}\|$, such as the Gaussian kernel:
\begin{equation}
	\kappa(\mr{i},\mr{j}) = \exp\left( -\frac{1}{2s^2} \|\mr{i} -\mr{j}\|^2 \right)
\end{equation}
where $s>0$ is the kernel bandwidth. In the sequel, we shall use the Gaussian kernel for its smoothness and non-informativeness, as we lack any knowledge about the unknown function $\psi$. Note that this kernel has been used successfully in many signal and image processing applications, in particular for hyperspectral data unmixing~\cite{chen2013nonlinear,chen2014nonlinear2}.

\section{Detection of nonlinearly mixed pixels}
\label{sec:Detection}

\subsection{The detection problem}

Given an observation $\br$, we formulate the nonlinear mixture detector as the following binary hypothesis test problem
\begin{subequations}
\begin{align}
	\label{eq:H0}
	{\cp H}_0&: \br = \MM\balpha + \bn\\
	\label{eq:H1}
	{\cp H}_1&: \br = \cb\psi(\MM) + \bn
\end{align}
\end{subequations}
where $\bn$ is a zero-mean WGN with variance $\sigma_n^2$.  We assume that the endmember matrix $\MM$ is available, or has been estimated from data using an endmember extraction technique~\cite{Bioucas-Dias-2013-ID307}. We shall relax this hypothesis in Section~\ref{sec:EMExtraction}, and use the nonlinear mixture detector to jointly perform this task.

We propose to compare the fitting errors resulting from estimating $\br$ with a linear or a nonlinear estimator~\eqref{eq:GPREst}. Under  ${\cp H}_0$, both estimators should provide good estimates. Under ${\cp H}_1$, the estimation error resulting from the linear estimator should be significantly larger than that obtained with the nonlinear estimator. We shall now evaluate these fitting errors.

\subsection{Linear estimation error}

The MMSE estimator~\eqref{eq:GPREst} may be used with the linear kernel~\eqref{eq:linear-kernel} to estimate $\balpha$ in \eqref{eq:H0}. Nevertheless, this would require to solve~\eqref{eq:hyperparameter} in order to estimate $\sigma_n^2$. To save on unnecessary computational efforts, we shall limit the use of GP to nonlinear model estimation. The MMSE estimator for \eqref{eq:H0} is given by:
\begin{equation}
	\label{eq:ls_estim}
	\hat{\balpha} = (\MM^{\top}\MM)^{-1}\MM^{\top} \br
\end{equation}
resulting in the following estimation error:  
\begin{equation}
	\be_\text{lin} = \br - \hat{\br}_\text{lin} = \cb{P} \br
\end{equation}
where $\cb{P}= \bI_L - \MM (\MM^{\top} \MM)^{-1}\MM^{\top}$ is an $L\times L$ projection matrix of rank $\rho=L-R$. Note that no constraint is imposed on the abundance vector $\balpha$. The objective is to obtain the best linear estimator, since the purpose at this point is not to perform unmixing, but to decide on the linearity (or not) of the considered mixture.

Consider first the distribution for $\|\cb{e}_\text{lin}\|^2$. Under $\cp{H}_1$, we have:
\begin{equation}
	\cb{e}_\text{lin}| {\cp{H}}_1 = \cb{P} [\cb{\psi} + \bn].
\end{equation}
This implies that
\begin{equation}
	\cb{e}_\text{lin}| {\cp{H}}_1 \sim \mathcal{N} (\cb{P}\cb{\psi}, \sigma_n^2 \cb{P})
 	\label{eq:eH1CondDist}
\end{equation}
where we use that the projection matrix $\cb{P}$ is idempotent, that is, $\sigma_n^2\cb{P}\cb{P}^\top = \sigma_n^2 \cb{P}$. Under ${\cp{H}}_0$, we have: 
\begin{equation}
	\cb{e}_\text{lin} | {\cp{H}}_0 \sim \mathcal{N} (0, \sigma_n^2\cb{P}).
\end{equation}
Proper normalization of each squared entry $e_{\text{lin},i}$ of $\cb{e}_\text{lin}$ leads to the following conditional distributions under the two hypotheses:
\begin{equation}
	\label{eq:el}
	\begin{split}
	\left. \frac{e_{\text{lin},i}^2}{\sigma_n^2\,\cb{p}^\top_i\cb{p}_i} \right | {\cp{H}}_1 
	&\sim \chi_1^2\left(\frac{[\cb{p}^\top_i\cb{\psi}]^2}{\sigma^2_n\,\cb{p}^\top_i\cb{p}_i}\right) \\
	\left. \frac{e_{\text{lin},i}^2}{\sigma^2_n\,\cb{p}^\top_i\cb{p}_i} \right | {\cp{H}}_0& \sim \chi_1^2\left(0\right)
	\end{split}
\end{equation}
where $\bp^\top_i$ denotes the $i$-th row of matrix $\cb{P}$, and  $\chi^2_n(\lambda)$ is the noncentral $\chi$-square distribution with $n$ degrees of freedom and centrality parameter~$\lambda$~\cite{PapoulisBook}.
As $\cb{P}$ is idempotent and of rank $\rho=L-R$, which leads to $\|\cb{e}_\text{lin}\|^2=\cb{r}^\top\cb{P}\cb{r}$, we conclude that~\cite[p. 33]{kay1998fundamentals}:
\begin{equation}
	\label{eq:chi-H0}
	\left. \frac{\|\cb{e}_\text{lin}\|^2}{\sigma_n^2} \right | {\cp{H}_0} \sim \chi^2_{\rho}\left(0\right).
\end{equation}

\subsection{Nonlinear estimation error with GP}

Since our interest at this point is not to make predictions for new data, but to evaluate the fitting error between the model output and the available data, we define the GP estimation error as:
\begin{equation}
	\label{eq:GPError}
	\be_\text{nlin} = \br - \hat{\br}_\text{nlin}
\end{equation}
where $\hat{\br}_\text{nlin}$ is given by \eqref{eq:GPREst} with $\MM_* = \MM$. Hence, using~\eqref{eq:GPREst} in~\eqref{eq:GPError} yields
\begin{equation}
	\label{eq:GP_residual}
	\cb{e}_\text{nlin} = \br -\widehat{\cb{\psi}}_*\Big|_{\MM_* = \MM} = \cb{H}\br 
\end{equation}
where $\cb{H} = \bI_L - \bK^{\top} \left[\bK +\sigma_{n}^2\bI_L\right]^{-1}$ is a real-valued matrix of rank $L$.

We shall now analyze the distribution of $\|\cb{e}_\text{nlin}\|^2$ under hypotheses ${\cp{H}}_0$ and ${\cp{H}}_1$. Under hypothesis ${\cp{H}}_1$, we have:
\begin{equation}
	\cb{e}_\text{nlin}| {\cp{H}}_1 = \cb{H}(\cb{\psi} +\bn).
\end{equation}
This leads to the following conditional distribution
\begin{equation}
	\cb{e}_\text{nlin}| {\cp{H}}_1 \sim \mathcal{N}( \cb{H}\cb{\psi}, \sigma_n^2 \cb{H}\cb{H}^\top).
\end{equation}
Under hypothesis ${\cp{H}}_0$, the distribution for the error becomes
\begin{equation}
	\label{eq:EgUnderHo}
	\cb{e}_\text{nlin} | {\cp{H}}_0 \sim \mathcal{N}( \cb{H}\MM\balpha, \sigma_n^2 \cb{H}\cb{H}^\top).
\end{equation}
The distribution of the $i$-th entry of $\cb{e}_\text{nlin}$ is thus given by
\begin{equation}
	e_{\text{nlin},i} | {\cp{H}}_0 \sim \mathcal{N}( \cb{h}^\top_i\MM\balpha, \sigma_n^2 \cb{h}^\top_i\cb{h}_i ).
\end{equation}
Proper normalization of each squared entry $e_{\text{nlin},i}$ of $\cb{e}_\text{nlin}$ yields the following conditional distributions:
\begin{equation}
	\label{eq:enl}
	\begin{split}
	\left. \frac{e_{\text{nlin},i}^2}{\sigma_n^2\,\cb{h}^\top_i\cb{h}_i} \right | {\cp{H}}_1 
	&\sim \chi_1^2\left(\frac{[\cb{h}^\top_i\cb{\psi}]^2}{\sigma^2_n\,\cb{h}^\top_i\cb{h}_i}\right) \\
	\left. \frac{e_{\text{nlin},i}^2}{\sigma^2_n\,\cb{h}^\top_i\cb{h}_i} \right | {\cp{H}}_0
	& \sim \chi_1^2\left(\frac{[\cb{h}^\top_i\cb{M\alpha}]^2}{\sigma^2_n\,\cb{h}^\top_i\cb{h}_i}\right)
	\end{split}
\end{equation}
where $\cb{h}^\top_i$ denotes the $i$-th row of $\cb{H}$. Non-central $\chi$-square distributions in~\eqref{eq:el} and \eqref{eq:enl} make the analysis of the test statistics in the next section intractable, even under $\cp{H}_0$. In order to proceed, we argue that it is reasonable to assume that, under $\cp{H}_0$, both the nonlinear GP regression method and the linear one should achieve the same level of accuracy. Considering \eqref{eq:chi-H0}, this approximation leads to 
\begin{equation}
	\label{eq:chi-H1}
	\left. \frac{\|\cb{e}_\text{nlin}\|^2}{\sigma_n^2} \right | {\cp{H}_0} = \chi^2_{\rho}(0).
\end{equation}
We validated this approximation using extensive Monte Carlo simulations. Figures~\ref{fig:Hists}a and \ref{fig:Hists}b illustrate this assumption for a representative example.

\subsection{The test statistics}
\label{sec:T}

We propose to compare the squared norms of the two fitting error vectors $\cb{e}_\text{nlin}$ and $\cb{e}_\text{lin}$ to decide between ${\cp H}_0$ and ${\cp H}_1$. Also, the test statistics should allow for the adjustment of the detection threshold to a given probability of false alarm (PFA) for design purposes.  Considering these two objectives, we propose the following statistical test
\begin{equation}
	T = \frac{2\|\cb{e}_\text{nlin}\|^2}{\|\cb{e}_\text{nlin}\|^2 +  
	\|\cb{e}_\text{lin}\|^2} \overset{\mathcal{H}_1}{\underset{\mathcal{H}_0}{\lessgtr}} \tau
 \label{eq:AllSampleDetector}
\end{equation}
where $\tau$ is the detection threshold.  

The reasoning behind the choice of $T$ defined in \eqref{eq:AllSampleDetector} is as follows. Under ${\cp H}_0$, both $\|\cb{e}_\text{nlin}\|^2$ and $ \|\cb{e}_\text{lin}\|^2$ are $\chi$-square dependent random variables. Now, we write $ \cb{e}_\text{lin}$ as $\cb{e}_\text{nlin} + \sqrt{2}\cb{\epsilon}$, where $\cb{\epsilon}$ is assumed to be also a zero-mean i.i.d. Gaussian vector\footnote{The constant factor $\sqrt{2}$ is  for notation purpose only.}, and neglect the cross-term $\cb{e}_\text{nlin}^{\top}\cb{\epsilon}$ when compared to $\|\cb{\epsilon}\|^2$ in evaluating 
$ \|\cb{e}_\text{lin}\|^2$ under ${\cp H}_0$. The latter approximation is due to the lack of correlation between $\cb{e}_\text{nlin}$ and $\cb{\epsilon}$, which can be largely attributed to  mismatches resulting from the numerical optimization required to solve  \eqref{eq:hyperparameter}. Under these considerations, \eqref{eq:AllSampleDetector} can be written as $T = \|\cb{e}_\text{nlin}\|^2/(\|\cb{e}_\text{nlin}\|^2 + \|\cb{\epsilon}\|^2)$ with both $\|\cb{e}_\text{nlin}\|^2$ and $\|\cb{\epsilon}\|^2$ independent and $\chi$-square distributed.  Such a statistics is known to follow a beta distribution~\cite{Johnson1995}.

As the GP estimator tends to fit better nonlinearly mixed data, $T$ should be less than 1 under hypothesis ${\cp H}_1$.  Conversely,  $T$ should be close to one for linearly mixed pixels, as $\|\cb{\epsilon}\|^2$ tends to be much less than $2\|\cb{e}_\text{nlin}\|^2$. Hence, as per \eqref{eq:AllSampleDetector}, we accept hypothesis ${\cp H}_0$ if $T>\tau$ and we conclude for the nonlinear mixing hypothesis ${\cp H}_1$ if $T<\tau$.

\subsection{Determining the detection threshold}

Considering the assumption that the statistical test $T$ has a beta distribution under $\cp{H}_0$, a decision threshold $\tau$ can be determined for a given PFA as
\begin{equation}
	\tau = \cp{B}_{\alpha,\beta}^{-1}(\text{PFA})
 \label{eq:tauDet}
\end{equation}
where $\cp{B}_{\alpha,\beta}$ is the cumulative distribution function of the beta distribution with parameters $(\alpha,\beta)$. The parameters of this function must be estimated from the data. To this end, we initially determine an estimate $\widehat{\!\cb{A}}$ of the abundance matrix assuming the linear mixing model with the real observations $\cb{R}=[\br_1,\ldots,\br_N]$ and the known endmember matrix $\MM$.  Then, using $\MM$ and $\widehat{\!\cb{A}}$ we construct the synthetic image $\cb{R}_{s} = \MM\widehat{\!\boldsymbol{A}}$, which satisfies $\cp{H}_0$.  For this linearly mixed hyperspectral image, we then compute, say, $N$ samples of the test statistics $T|{\cp H}_0$ defined in~\eqref{eq:AllSampleDetector} and fit a beta distribution to these samples.  The threshold $\tau$ for each PFA is then determined using \eqref{eq:tauDet}.

This procedure requires the knowledge of the endmember matrix $\MM$.  The next section proposes an iterative technique to estimate $\MM$ from an hyperspectral image, which we assume to contain linearly and nonlinearly mixed pixels.

\section{Endmember extraction in nonlinearly mixed hyperspectral images}
\label{sec:EMExtraction}

The presence of nonlinearly mixed pixels in an hyperspectral image tends to degrade the estimation accuracy of endmember extraction methods based on a linear mixing model. As a consequence, nonlinearly mixed pixels also affect the performance of algorithms using the endmember matrix such as the detection method presented in this paper. There has been few papers addressing endmember estimation from nonlinearly mixed images. A nonlinear unmixing algorithm is derived in~\cite{altmann2013unsupervised}. The pixel reflectances are supposed to be post-nonlinear functions of unknown pure spectral components. A Bayesian strategy is proposed to both unmix the data and estimate the endmembers. Both tasks are however mutually dependent and the unmixing model is very specific.  A nonlinear endmember estimation algorithm based on the approximation of geodesic distances is introduced in~\cite{Heylen2011nonlinear,Nguyen2012}. This algorithm can however suffer from the absence of pure pixels in the image, and the effectiveness of using manifold learning methods on real data still needs to be analyzed and confirmed. In this section, we propose an iterative technique for estimating the endmember matrix~$\MM$ under the reasonable assumptions that the number $R$ of endmembers is known~\cite{chang2004estimation,bioucas2008hyperspectral,halimi2015estimating}, and that these endmembers are linearly mixed within at least a small part of the image. Nonlinear mixtures may however compose a significant part of the image.  The proposed technique combines the detector of nonlinearly mixed pixels presented in Section~\ref{sec:Detection} and the endmember estimation algorithm MVES~\cite{Chan-2009-ID334}. 

The procedure is described in Algorithm~\ref{alg:MVIEE}. It is a two-step iterative algorithm. The first step consists of using MVES to estimate the endmembers (line 2 and 14 in Algorithm~\ref{alg:MVIEE}). The second step uses~\eqref{eq:AllSampleDetector} to compute the detection statistics for all the $L$ pixels in the image $\bR_{\text{tmp}}$ (line 7 in Algorithm~\ref{alg:MVIEE}). Then, all pixels whose detection statistic satisfies ${T}(i)\leq \tau_{r}$ are removed (line 9), where $\tau_{r}=r_{f}\times \tau$ (line 4 and 11) is the relaxed detection threshold. The use of a relaxed threshold is suggested to avoid discarding linear pixels during the first iterations, when the estimates of $\MM$ are still not sufficiently accurate. The relaxing factor is initialized for $r_{f}=0.9$ and  is increased by a factor $r_{\rm inc} = 0.1/N_{\max}$ at each iteration to improve pixel selection as the estimation of the matrix $\MM$ improves (line 10). The procedure is repeated until the linear and the nonlinear GP models in~\eqref{eq:AllSampleDetector} present similar fitting errors within the limit of $\varepsilon$. Using this procedure, $\tau_{r}$ tends to the desired threshold $\tau$ as the estimation of $\MM$ improves, leaving mostly linear pixels for which both models have similar performance. A maximum number of iterations $N_{\max}$ is also set to avoid discarding too much data. 

Note that we have opted for the MVES algorithm for endmember extraction because it inscribes the data into a minimum-volume simplex. Thus, MVES is suitable to estimate $\MM$ in the absence of pure pixels. This feature is specially interesting for our purpose since the procedure described above discards data, which may even be pure or near-pure pixels during the first iterations. Nevertheless, any other endmember estimation algorithm valid in absence of pure pixel could be potentially used with Algorithm~\ref{alg:MVIEE}.

\begin{algorithm}
\SetKwInOut{Input}{Input}
\SetKwInOut{Output}{Output}
\caption{Iterative endmember estimation~\label{alg:MVIEE}}
 \Input{The hyperspectral image $\bR$, and the number of endmembers $R$}
 \Output{Estimated endmember matrix $\widehat{\MM}$}
 Initialization:
 $T_{\max}\!=1$, $T_{\min}\!=0$, $\varepsilon = 0.05$, $\bR_{\text{tmp}} = \bR$, $N_{\max} = 10$, $cc=1$, $r_f=0.9$, $r_{inc}=0.1/N_{\max}$, $\text{PFA}=0.05$\;
 $\widehat{\MM} = \text{MVES}(\bR_{\text{tmp}},R)$\;
Compute $\tau$ using~\eqref{eq:tauDet}\;
$\tau_{r} = r_f \times\tau$;\qquad\qquad\,\,\, \%\% (relaxed threshold)\

 \While {$ T_{\max} - T_{\min} > \varepsilon$ \& $cc <N_{\max}$}{
  \For{$i=1$ \KwTo $N_{\text{pixels}}$}{ 
    Compute $\cb{T}(i)$ using~\eqref{eq:AllSampleDetector}\;    
  }
  Remove all pixels with $\cb{T}(i)\leq\tau_{r}$\ from $\bR_{\text{tmp}}$\; 
  $r_f = r_f + r_{inc}$;\qquad \%\% (relaxing factor)\
  
  $\tau_r = r_f\times\tau$\;
  $T_{\max}\!=\max(\cb{T})$; $T_{\min}\!=\min(\cb{T})$\;
  $cc =cc + 1$\;
  $\widehat{\MM} = \text{MVES}(\bR_{\text{tmp}},R)$\;
 }
 
\end{algorithm}

\section{Simulations}
\label{sec:Sim}

This section presents simulation results to validate the proposed approach for detecting nonlinearly mixed pixels, with both synthetic and real images.  The use of synthetic images is important as they provide a ground truth against which the performance of the detector can be verified.  First, we propose a definition for a degree of nonlinearity of an hyperspectral image so that the relative performances of different detectors can be compared.  This is helpful to quantify the relative energies associated with the linear and nonlinear mixing components in hyperspectral images generated with different nonlinear mixing models.

\subsection{Degree of nonlinearity}
\label{sec:NLDegree}

Consider that a pixel vector can be written as the sum of a linear and a nonlinear mixing component\footnote{We do not account for noise contribution as it can be set by the user independently of the mixing model.} as is the case for most existing nonlinear mixing models~\cite{fan2009, Nascimento2009,Somers:2009p6577,altmann2011:icassp,halimi2011}:
\begin{equation}
	\br = \br_\text{lin}+ \br_\text{nlin}
 \label{eq:pixLNLmodel}
\end{equation}
where $\br_\text{lin}$ and $\br_\text{nlin}$ are, respectively, the linear and nonlinear mixing contributions to $\br$. The energy of $\br$ is given by
\begin{equation}
	E = \|\br\|^2 = \|\br_\text{lin}\|^2 + 2\,\br_\text{lin}^\top\br_\text{nlin} + \|\br_\text{nlin}\|^2,
\end{equation}
where $E_\text{lin} = \|\br_\text{lin}\|^2$ is the energy of the linear contribution and $E_\text{nlin} = 2\,\br_\text{lin}^\top\br_\text{nlin} + \|\br_\text{nlin}\|^2$ is the part of the pixel energy affected by the nonlinear mixing.  Given a mixing model, we define the degree of nonlinearity $\eta_d$ as the ratio of the energy of the nonlinear contribution $E_\text{nlin}$ to the total energy $E$. Thus,
\begin{equation}
	\eta_d 	= \frac{E_\text{nlin}}{E} = \frac{1}{1 + A}
	\label{eq:DofNl}
\end{equation}
where $A = \|\br_\text{lin}\|^2/(2\,\br_\text{lin}^\top\br_\text{nlin} + \|\br_\text{nlin}\|^2)$.
Next, we show how to apply this definition for generating synthetic samples with two different mixing models.

\subsubsection{Synthetic data generation with GBM}

To be able to control the relative contributions of the linear and nonlinear mixing parts of the GBM model, we introduce a new scaling factor $k$ into the generalized bilinear model (GBM) used in~\cite{Altmann-2013-ID306}. For an endmember matrix $\MM$ and an abundance vector $\balpha$, we write the modified noiseless GBM model as 
\begin{equation}
	\br = k\MM \balpha + \gamma\bnu
	\label{eq:MGBM}
\end{equation}
where $0 \le k \le 1$, $\bnu = \sum_{i=1}^{R-1}\sum_{j=i+1}^{R}\alpha_i \alpha_j \bm_i\odot \bm_j$ is the nonlinear mixing term, $\gamma$ is the scaling parameter for the nonlinear contribution, and $\odot$ is the Hadamard product.  
The degree of nonlinearity is then 
\begin{equation}
	\begin{split}
	\eta_d &= \frac{2k\gamma(\bnu^\top\MM\balpha) + \gamma^2\|\bnu\|^2}
	{k^2\|\MM\balpha\|^2 + 2k\gamma (\bnu^\top\MM\balpha) + \gamma^2\|\bnu\|^2}\\
	&= \frac{1}{1 + A}
	\end{split}
	\label{eq:eta_d}
\end{equation}
with $A=k^2\|\MM\balpha\|^2 / (2k\gamma(\bnu^\top\MM\balpha) + \gamma^2\|\bnu\|^2)$. We have to determine the scaling factors $k$ and $\gamma$ so that the energy $E$ is independent of $\eta_d\geq0$. This condition can be expressed as $\|\MM\balpha\|^2 = k^2\|\MM\balpha\|^2 + 2k\gamma(\bnu^\top\MM\balpha) + \gamma^2\|\bnu\|^2$, leading to
\begin{equation}
	A = \frac{k^2}{1-k^2}
\end{equation}
or
\begin{equation}
	k = \sqrt{\frac{A}{1+A}}= \sqrt{1- \eta_d}.
 	\label{eq:k1}
\end{equation}
To obtain $\gamma$, note that the denominator of $A$ can be written as $\gamma^2 |\bnu\|^2 +2{k}\gamma(\bnu^\top\MM\balpha) = (1-k^2)\|\MM\balpha\|^2$. Since $\gamma$ must be positive, we have 
\begin{equation}
	\gamma = \frac{1}{2\|\bnu\|^2}\Big(-2{k}(\bnu^\top\MM\balpha)
 	+\sqrt{4k^2(\bnu^\top\MM\balpha)^2+ 4\|\bnu\|^2(1-k^2)\|\MM\balpha\|^2}\Big).
 	\label{eq:gamma}
\end{equation}
Once $k$ and $\gamma$ have been determined from $\eta_d$, we can generate data following the model in~\eqref{eq:MGBM}.

\subsection{Synthetic data generation with PNMM}

To match the noiseless PNMM model \eqref{eq:pnmm_chen} with the proposed formulation \eqref{eq:MGBM}, we complement it with a weighted linear mixture as follows:
\begin{equation}
	\br = k\MM\balpha + \gamma \bnu,
	\label{eq:pnmNewModel}
\end{equation}
where $\bnu=(\MM\balpha)^\xi$ denotes the exponential value $\xi$ applied to each entry of $\MM\balpha$. Model~\eqref{eq:pnmNewModel} reduces to \eqref{eq:pnmm_chen} for $k=0$ and $\gamma=1$.  Again, parameters $k$ and $\gamma$ are scaling factors that control the relative amounts of linear and nonlinear contributions given $\eta_d$.  As for the GBM, both can be set using \eqref{eq:k1} and \eqref{eq:gamma}.

\subsection{Simulations with known $\MM$}
\label{sec:simKnownM}

We now present simulations with synthetic data and a known endmember matrix $\MM$. These simulations allow us to assess the detector performance disregarding estimation errors for the endmembers. Hence, they illustrate the potential of the proposed detector.  To construct synthetic data, we used three materials ($R=3$) extracted from the spectral library of the software ENVI\texttrademark~\cite{ENVI}: green grass, olive green paint and galvanized steel metal. Each endmember $\bm_r$ has $L=826$ bands that were uniformly decimated by 3 to $L=275$ bands. 

To evaluate the performance of the proposed detector, we generated 8000 synthetic samples by mixing the three collected spectra. Among the 8000 pixels, 4000 were generated using the linear model in~\eqref{eq:linForm}, and 4000 using the modified generalized bilinear model in~\eqref{eq:MGBM}. A fixed abundance vector $\balpha = [0.6,\, 0.4,\, 0.1]^\top$ was used for all samples.  Nonlinearly mixed samples were generated using different degrees of nonlinearity $\eta_d\in\{0.3,0.5,0.8\}$ to test the detector under different conditions. The power of the additive Gaussian noise was set to $\sigma_n^2 = 0.001$, which corresponds to $\text{SNR}=21\text{dB}$.

\begin{figure}
 \psfrag{ROC Curve}{}
 \psfrag{Empirical ROCs for the GP Detector}{}
 \psfrag{Probability of Detection}{\hspace{1.2cm}PD}
 \psfrag{Probability of False Alarm}{\hspace{1.2cm}PFA}
        \centering
        \begin{subfigure}[b]{0.45\textwidth}
	       \psfrag{NLD = 0.3}{$\!\eta_d\!=\!0.3$}
	        \psfrag{NLD = 0.5}{$\!\eta_d\!=\!0.5$}
	        \psfrag{NLD = 0.8}{$\!\eta_d\!=\!0.8$}
                \includegraphics[width=\textwidth]{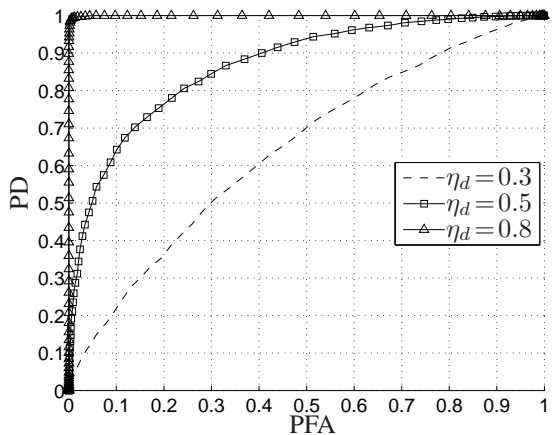}
                \caption{Robust LS detector~\cite{Altmann-2013-ID306}.}
                \label{fig:synthRes_A}
        \end{subfigure}%
        \\
        \begin{subfigure}[b]{0.45\textwidth}
		\psfrag{NLD = 0.3}{$\!\eta_d\!=\!0.3$}
	        \psfrag{NLD = 0.5}{$\!\eta_d\!=\!0.5$}
	        \psfrag{NLD = 0.8}{$\!\eta_d\!=\!0.8$}
                \includegraphics[width=\textwidth]{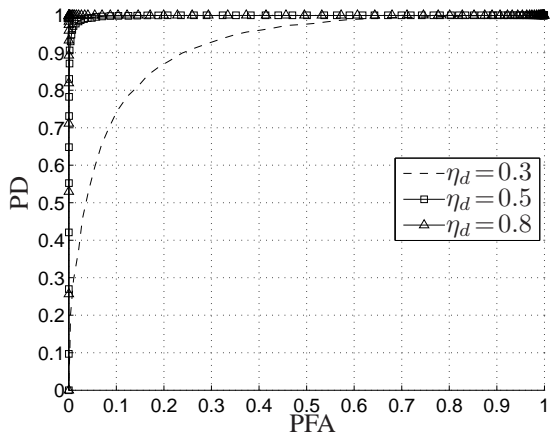}
                \caption{Proposed GP detector.}
                \label{fig:synthRes_B}
        \end{subfigure}
        \\        
        \begin{subfigure}[b]{0.45\textwidth}
                \includegraphics[width=\textwidth]{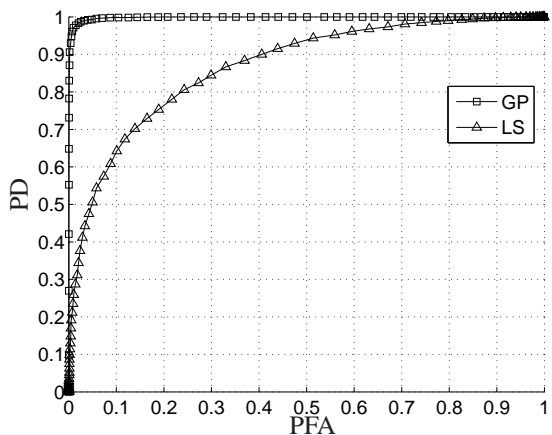}
                \caption{Comparison of the LS and GP detectors for $\eta_d=0.5$.}
                \label{fig:synthRes_C}
        \end{subfigure}
        \caption{Empirical ROCs for: (a) the Robust LS detector~\cite{Altmann-2013-ID306}, (b) the proposed GP detector, (c) the two detectors for $\eta_d = 0.5$. All curves were obtained for 8000 pixels (4000 linearly mixed and 4000 nonlinearly mixed) and $\text{SNR}=21$dB. Nonlinear mixtures were generated using the simplified GBM described in Section~\ref{sec:NLDegree}. }
\label{fig:synthRes}
\end{figure}

\begin{figure*}
        \centering
        \begin{subfigure}[b]{0.33\textwidth}
		\psfrag{Squared Norm of the GP Fitting Error under H0}{\footnotesize{\hspace{0.9cm}$\|\be_\text{nlin}\|^2$ under $\cp{H}_0$}}
		\psfrag{Squared Norm of the GP Fitting Error under H1}{\footnotesize{\hspace{0.9cm}$\|\be_\text{nlin}\|^2$ under $\cp{H}_1$}}
                \includegraphics[width=\textwidth]{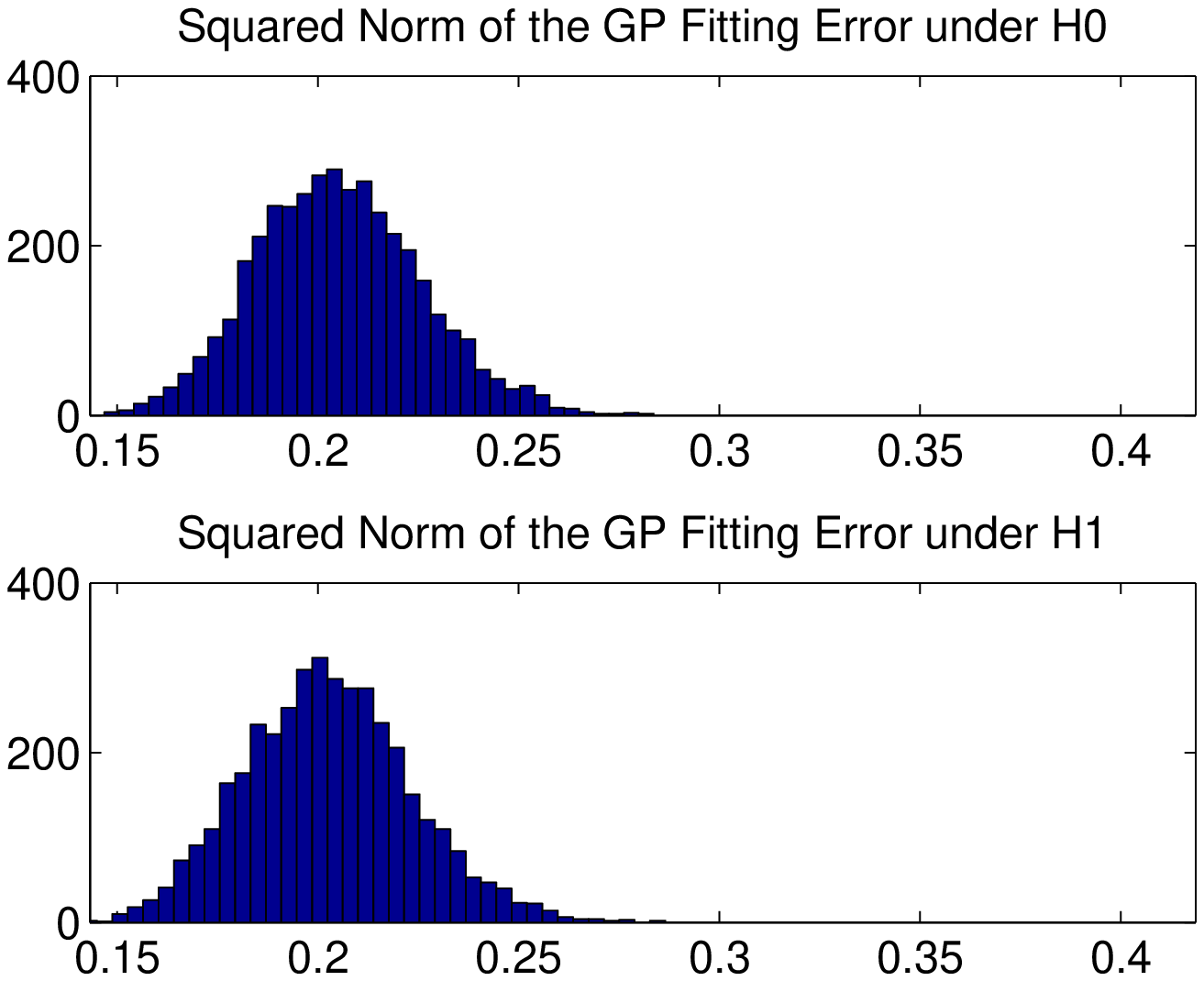}
                \caption{Histogram of $\|\be_g\|^2$.}
                \label{fig:Hists_A}
        \end{subfigure}%
        \begin{subfigure}[b]{0.33\textwidth}
        	\psfrag{Squared Norm of the LS Fitting Error under H0}{\footnotesize{\hspace{0.9cm}$\|\be_\text{lin}\|^2$ under $\cp{H}_0$}}
		\psfrag{Squared Norm of the LS Fitting Error under H1}{\footnotesize{\hspace{0.9cm}$\|\be_\text{lin}\|^2$ under $\cp{H}_1$}}
                \includegraphics[width=\textwidth]{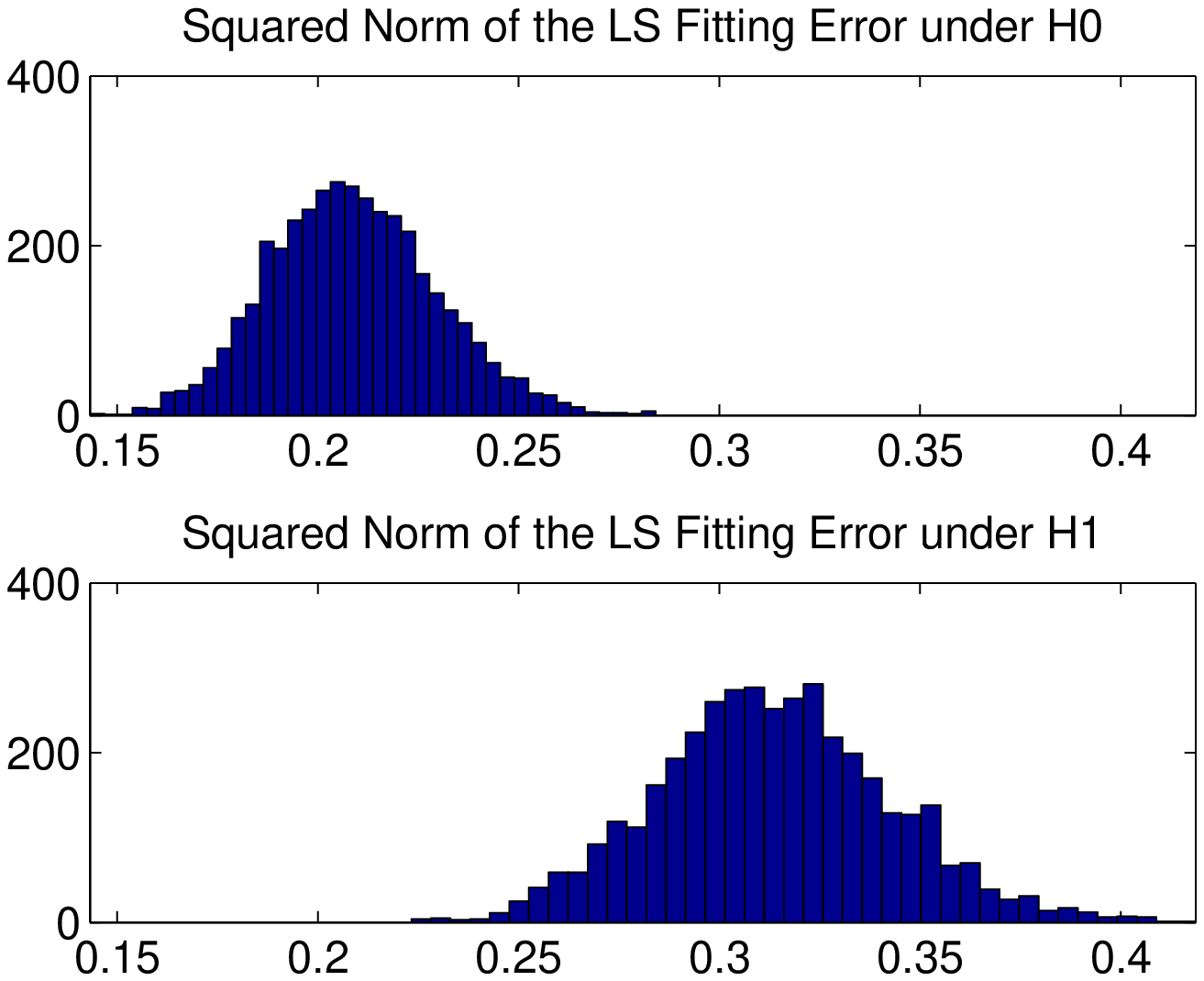}
                \caption{Histogram of $\|\be_l\|^2$.}
                \label{fig:Hists_B}
        \end{subfigure}
        \begin{subfigure}[b]{0.33\textwidth}
		\psfrag{Proposed Test Statistics under H0}{\footnotesize{\hspace{0.7cm}$T$ under $\cp{H}_0$}}
		\psfrag{Proposed Test Statistics under H1}{\footnotesize{\hspace{0.7cm}$T$ under $\cp{H}_1$}}
                \includegraphics[width=\textwidth]{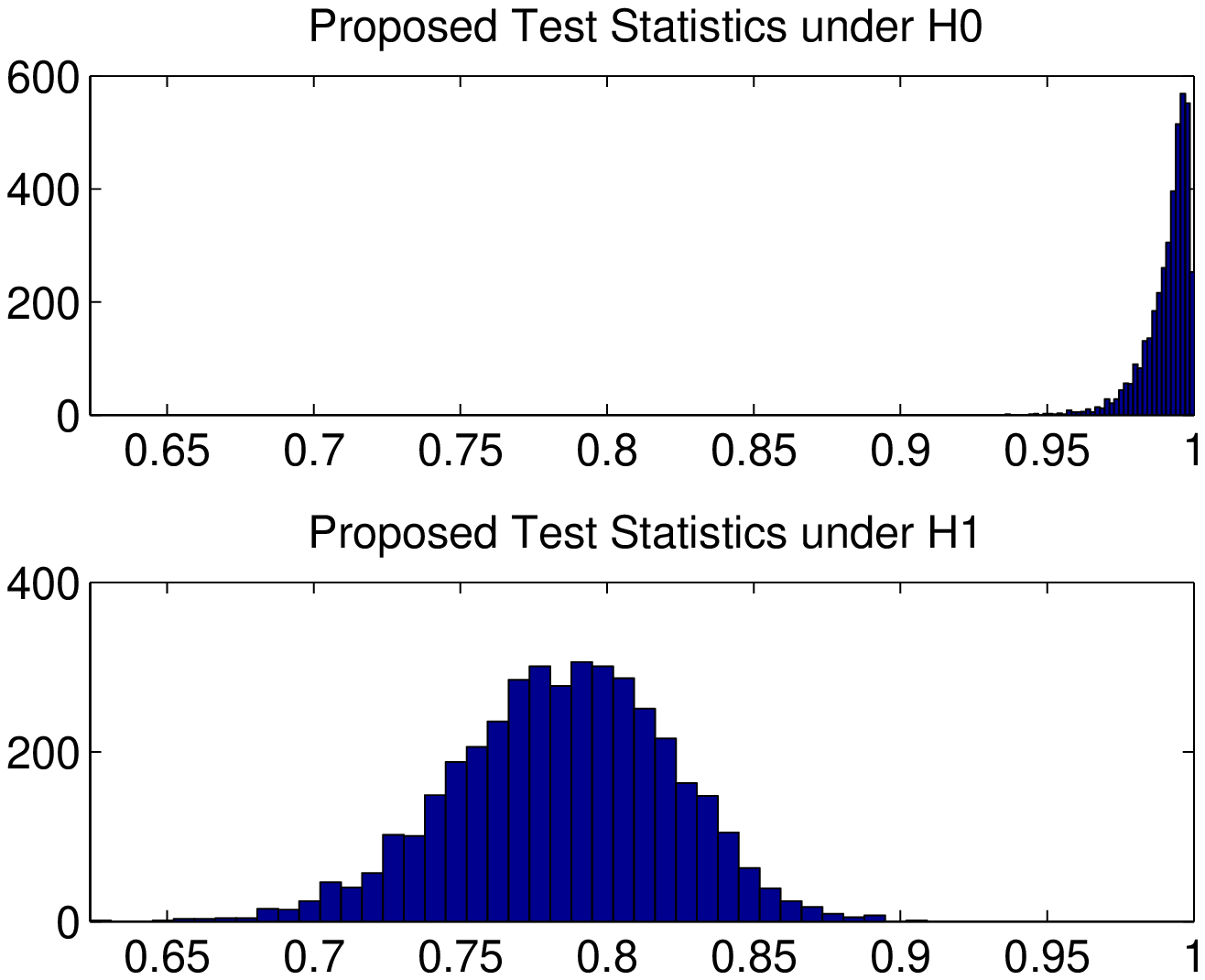}
                \caption{Histogram for the test statistics~\eqref{eq:AllSampleDetector}.}
                \label{fig:Hists_C}
        \end{subfigure}
        \caption{Histograms for (a) the squared norm of the GP fitting error, (b) the least-squares fitting error, and (c) the test statistics~\eqref{eq:AllSampleDetector}.}\label{fig:Hists}
\end{figure*}

\begin{figure}
 \centering
 \includegraphics[width=0.45\textwidth]{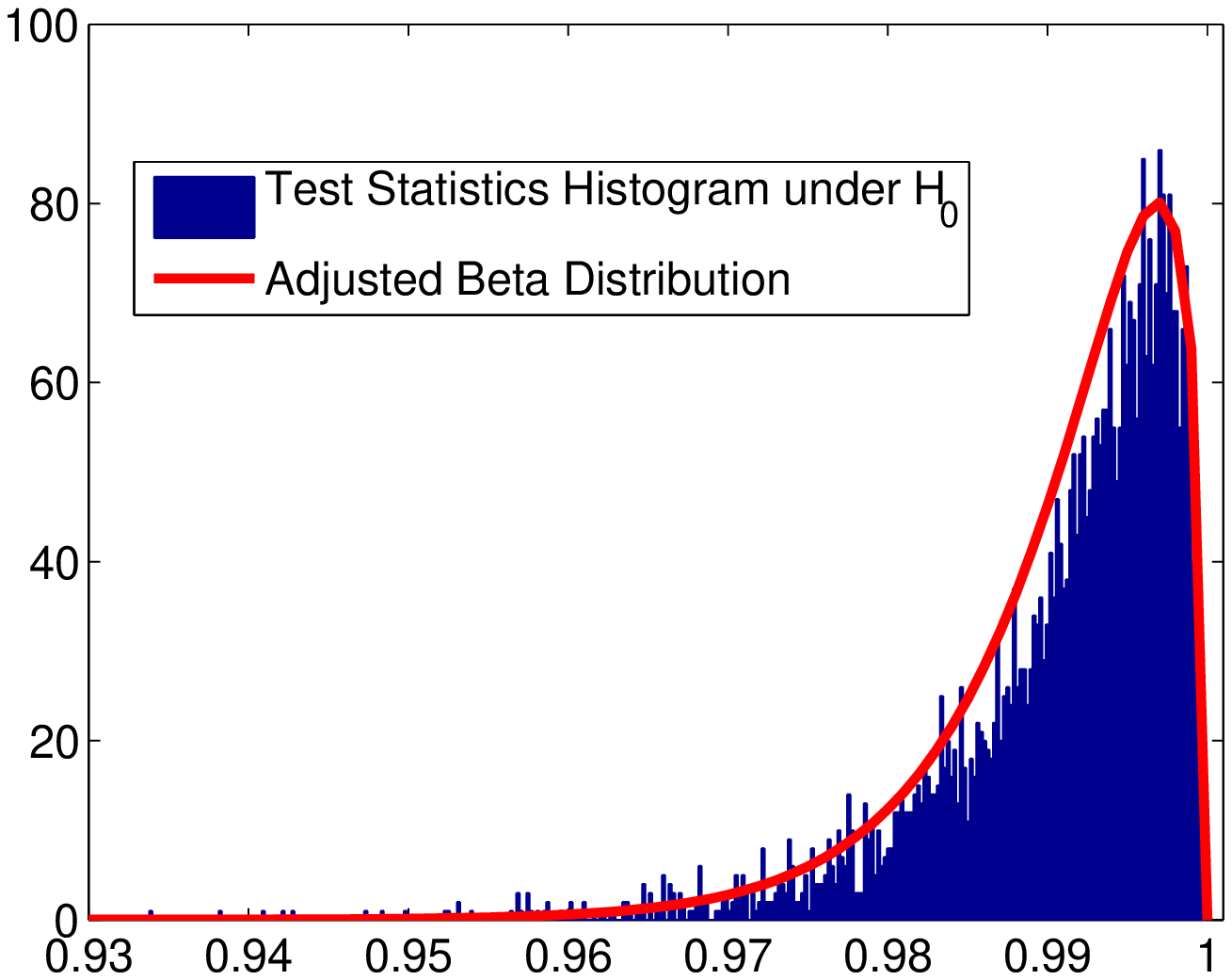}
 \caption{Histogram of the test statistics under $\cp{H}_0$ and the adjusted Beta distribution.}\label{fig:BetaPlot}
\end{figure}

Figure~\ref{fig:synthRes} shows the receiver operating characteristics (ROCs) of the proposed GP detector and the LS robust detector presented in~\cite{Altmann-2013-ID306} for the three values of $\eta_d$.  The proposed detector performs better, especially for moderate to high degree of nonlinearity.  For instance, Fig.~\ref{fig:synthRes}c shows that the GP detector achieves a probability of detection of $1$ for $\text{PFA}=0.1$, while the LS robust detector yields a probability of detection of approximately $0.65$ for the same PFA. Figure~\ref{fig:Hists} shows the histograms of $\|\be_\text{nlin}\|^2$, $\|\be_\text{lin}\|^2$ and $T$ for both linearly ($\cp{H}_0$) and nonlinearly ($\cp{H}_1$) mixed data. The proposed test statistics clearly leads to histograms that differ significantly under both hypotheses $\cp{H}_0$ and $\cp{H}_1$, which explains the improvement in detection performance. Figure~\ref{fig:BetaPlot} compares the histogram of $T$ under $\cp{H}_0$ with the fitted beta distribution, confirming that the distribution of $T$ can be reasonably approximated by a beta distribution.

%
%

We considered two unmixing algorithms to assess the impact of the proposed detector on unmixing performance, one linear and one nonlinear. Linear unmixing was performed using the fully-constrained least-squares (FCLS) algorithm~\cite{Heinz:2001ep}.  For nonlinear unmixing, we used the SK-Hype algorithm~\cite{chen2013nonlinear}. The two algorithms were employed in two unmixing strategies.  First, each algorithm was used to unmix the complete hyperspectral image.  In the second strategy called Detect-then-Unmix, the proposed detector was used as a pre-processing step.  Then, FCLS was used to unmix pixels detected as linearly mixed and SK-Hype was used to unmix pixels detected as nonlinearly mixed.  The detection threshold $\tau$ was determined for $\text{PFA}=0.01$.  Two synthetic images were considered with $1000$ pixels each, $500$ being linearly mixed and $500$ being nonlinearly mixed.  Each image was constructed using a particular nonlinear mixing model, with a fixed degree of nonlinearity $\eta_d = 0.5$ in both cases. The GBM~\eqref{eq:MGBM} was used for the first image, while the PNMM~\eqref{eq:pnmNewModel} with $\xi = 3$ was considered for the second image. The SNR was 21dB in both cases, and the abundances were drawn uniformly in the simplex. Parameters $k$ and $\gamma$ were determined for each pixel vector to maintain the desired value of nonlinearity degree $\eta_d$ for all simulations.  To compare the results, we used the {root mean square error} (RMSE) of abundance estimation, defined as
\begin{equation}
	\text{RMSE} = \sqrt{\frac{1}{NR}\sum_{n=1}^{N}\|\balpha_n-\hat{\balpha}_n\|^2}
	\label{eq:RMSEAlpha}
\end{equation}
where $N$ is the number of pixels in each image. 

The results are presented in Tables~\ref{tab:recErrorknownMGBM} and~\ref{tab:recErrorknownMPNMM}. For each image, these tables indicate the RMSEs for the linearly mixed part (LMM), for the nonlinearly mixed part (NLM), and for the full image (Full Img.) using the three unmixing strategies. The results in blue are those with the lowest RMSE in each row of the tables.  As expected, FCLS has the best results when unmixing linearly mixed pixels. The same observation can be made for SK-Hype with nonlinearly mixed pixels. Nevertheless, we verify that the results using the Detect-then-Unmix strategy are very close to the best results for both types of pixels, LMM and NLM. When processing the whole image without prior information on the mixing nature of each pixel, the best results were those obtained with the Detect-then-Unmix strategy. 
\begin{table*}
\caption{RMSE in abundance estimation for $\MM$ known and using the GBM mixing model (SNR = 21dB, $\eta_d=0.5$).}\label{tab:recErrorknownMGBM}
 \centering
 {\renewcommand{\arraystretch}{1.2}
\begin{tabular}{|c|c|c|c|}
\cline{2-4}
\multicolumn{1}{c|}{}&\multicolumn{3}{|c|}{Image I: LMM + GBM} \\\hline
Model & FCLS & SK-Hype & Detect-then-Unmix \\ \hline
LMM & \cblue{0.0173 $\pm$ 3.20e-04} & 0.0349 $\pm$ 0.001543 & 0.0189 $\pm$ 6.44e-04 \\ 
NLM & 0.1220 $\pm$ 0.016268 & \cblue{0.0644 $\pm$ 0.003463} & 0.0666 $\pm$ 0.004054 \\ 
Full Img.& 0.0871 $\pm$ 0.01362 & 0.0518 $\pm$ 0.003054 & \cblue{0.0490 $\pm$ 0.003548}\\\hline
\end{tabular}}
\end{table*}

\begin{table*}
\caption{RMSE in abundance estimation for $\MM$ known and using the PNMM mixing model (SNR = 21dB, $\eta_d=0.5$).}\label{tab:recErrorknownMPNMM}
 \centering
 {\renewcommand{\arraystretch}{1.2}
\begin{tabular}{|c|c|c|c|}
\cline{2-4}
\multicolumn{1}{c|}{}&\multicolumn{3}{|c|}{Image II: LMM + PNMM}\\\hline
Model & FCLS & SK-Hype & Detect-then-Unmix\\ \hline
LMM & \cblue{0.0170 $\pm$ 3.14e-04} & 0.0354 $\pm$ 0.001723 & 0.0173 $\pm$ 3.76e-04\\ 
NLM & 0.0637 $\pm$ 0.003696 & \cblue{0.0551 $\pm$ 0.003745} & 0.0559 $\pm$ 0.003636\\ 
Full Img.& 0.0466 $\pm$ 0.003228 & 0.0463 $\pm$ 0.003047 & \cblue{0.0414 $\pm$ 0.002946}\\\hline
\end{tabular}}
\end{table*}


\subsection{Simulations with an unknown endmember matrix $\MM$}
\label{sec:simUnknownM}

The simulations conducted in Section~\ref{sec:simKnownM} assumed the endmember matrix $\MM$ to be known. Although this study is important to quantify the potential of the proposed detector, the endmembers are rarely known in practice.  Hence, in this section, we study the sensitivity of the detection performance as a function of the endmember estimation accuracy and of the degree of nonlinearity. Endmember extraction was performed with the iterative method proposed in Section~\ref{sec:EMExtraction}, and with VCA \cite{Nascimento2005} for comparison. 

\begin{figure}
  \centering
  \includegraphics[width=0.45\textwidth]{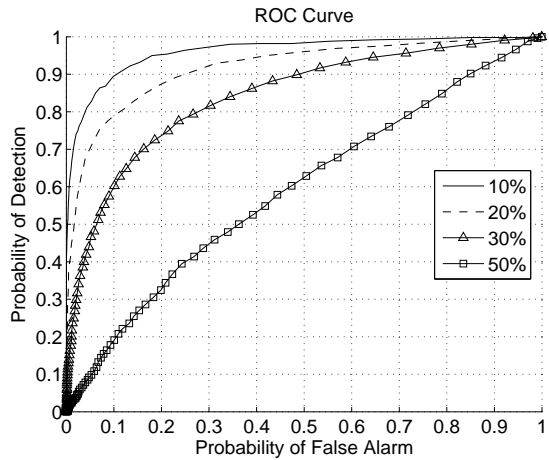}
  \vspace{-2mm}
\caption{ROCs for different proportions of nonlinearly mixed pixels and $\eta_d=0.5$. Endmember extraction using VCA.}
\label{fig:detThr_A}
\end{figure}

\begin{figure}
 \centering
\psfrag{NLD=0.3}{$\!\eta_d\!=\!0.3$}
\psfrag{NLD=0.5}{$\!\eta_d\!=\!0.5$}
\psfrag{NLD=0.8}{$\!\eta_d\!=\!0.8$}
 \includegraphics[width=0.48\textwidth]{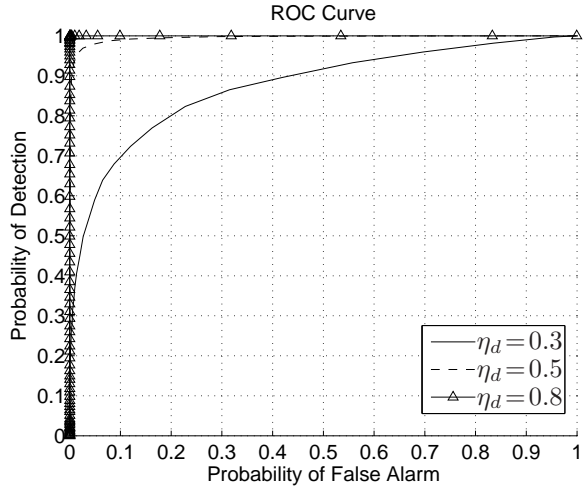}
 \caption{ROCs for different degrees of nonlinearity  $\eta_d$ and 50\% of nonlinearly mixed pixels in the image. Endmember extraction using Algorithm~\ref{alg:MVIEE}}\label{fig:iterativeROCs}
\end{figure}

Figure~\ref{fig:detThr_A} presents the results of $4$ experiments using synthetic images with $5000$ samples, $\text{SNR}=21\text{dB}$, abundances uniformly sampled in the simplex, a proportion of nonlinearly mixed pixels in the image varying from 10\% to 50\%, and $\eta_d=0.5$. For every experiment, the endmember matrix was extracted using VCA. These results show how the detection performance can degrade as the number of nonlinear pixels increases and as VCA loses accuracy in extracting the endmembers from the image. These results confirm the importance of devising alternatives to VCA (or tp other endmember extraction algorithms specifically designed for linearly-mixed images) for images containing nonlinearly-mixed pixels. Figure~\ref{fig:iterativeROCs} presents the results obtained with Algorithm~\ref{alg:MVIEE} for endmember extraction. For this experiment, we generated data with 50\% of nonlinearly mixed pixels and different degrees of nonlinearity $\eta_d\in\{0.3, 0.5, 0.8\}$. Comparing Fig.~\ref{fig:synthRes} and~\ref{fig:iterativeROCs} shows that the results obtained with the iterative endmember extraction algorithm are very close to those obtained for a known endmember matrix $\MM$ (which can be considered as the reference detector). 

\begin{figure*}
        \centering
        \begin{subfigure}[b]{0.45\textwidth}
                \includegraphics[width=\textwidth]{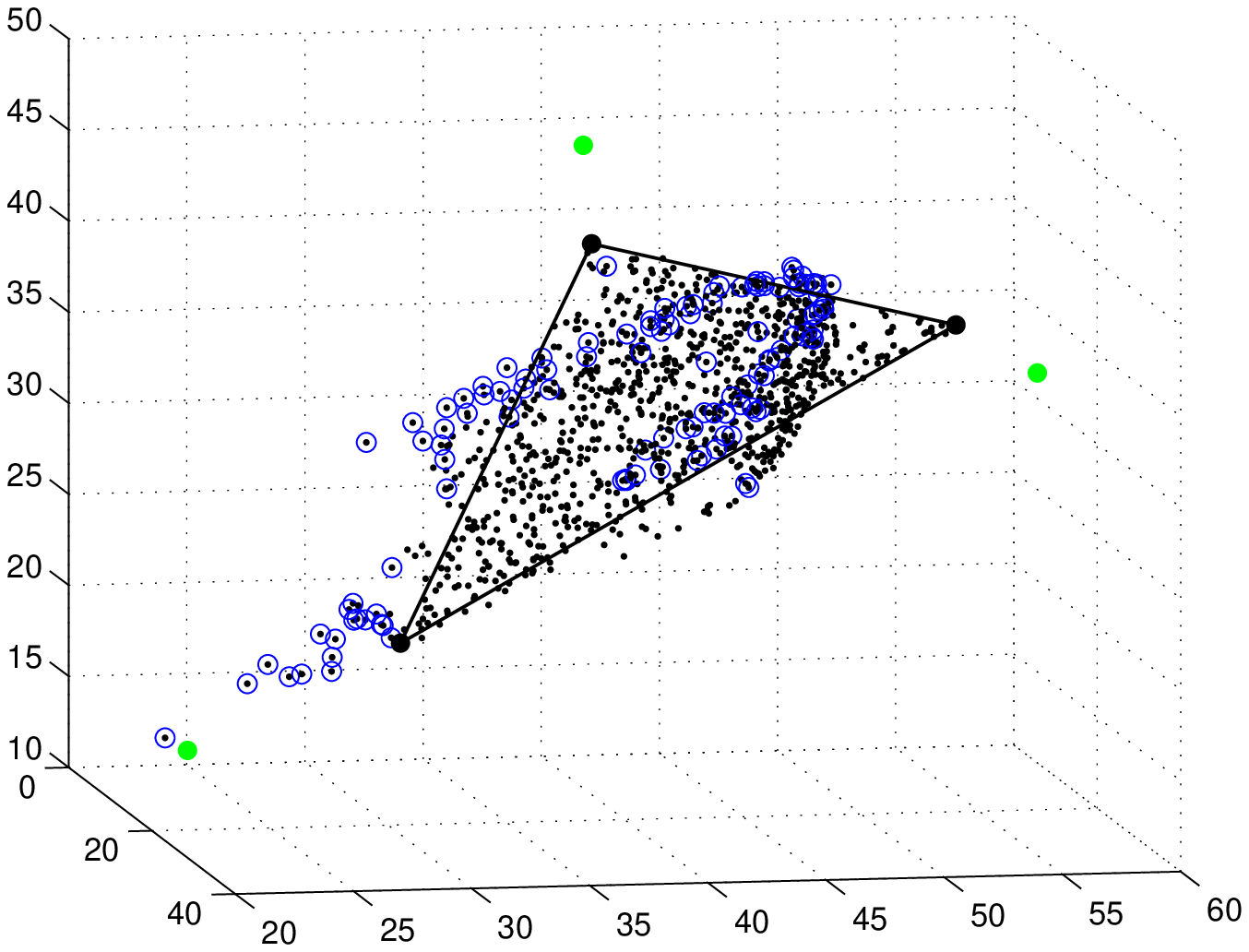}
                \caption{First iteration.}
                \label{fig:itt1}
        \end{subfigure}%
        ~
        \begin{subfigure}[b]{0.45\textwidth}
                \includegraphics[width=\textwidth]{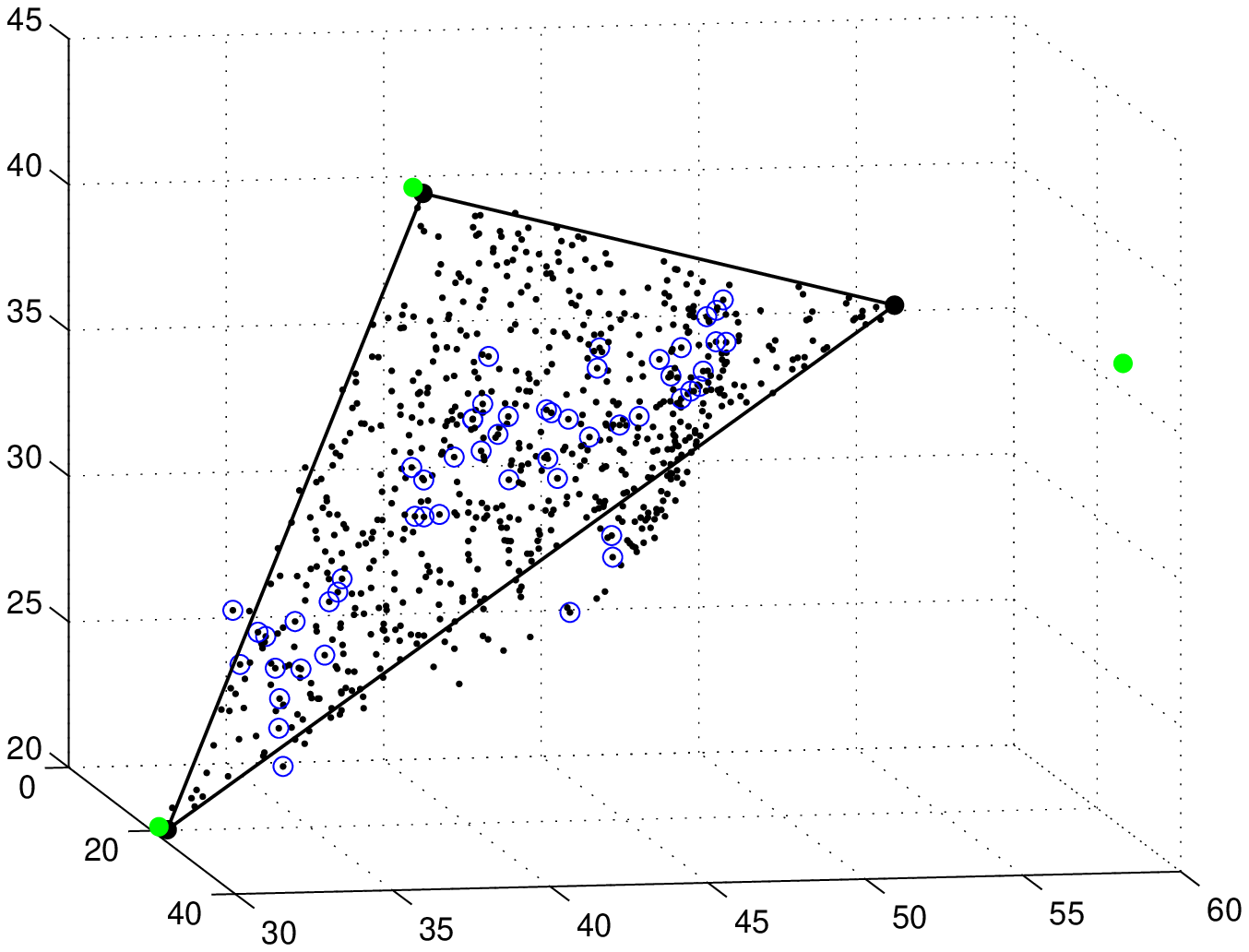}
                \caption{Fourth iteration.}
                \label{fig:itt4}
        \end{subfigure}
        \\
        \begin{subfigure}[b]{0.45\textwidth}
                \includegraphics[width=\textwidth]{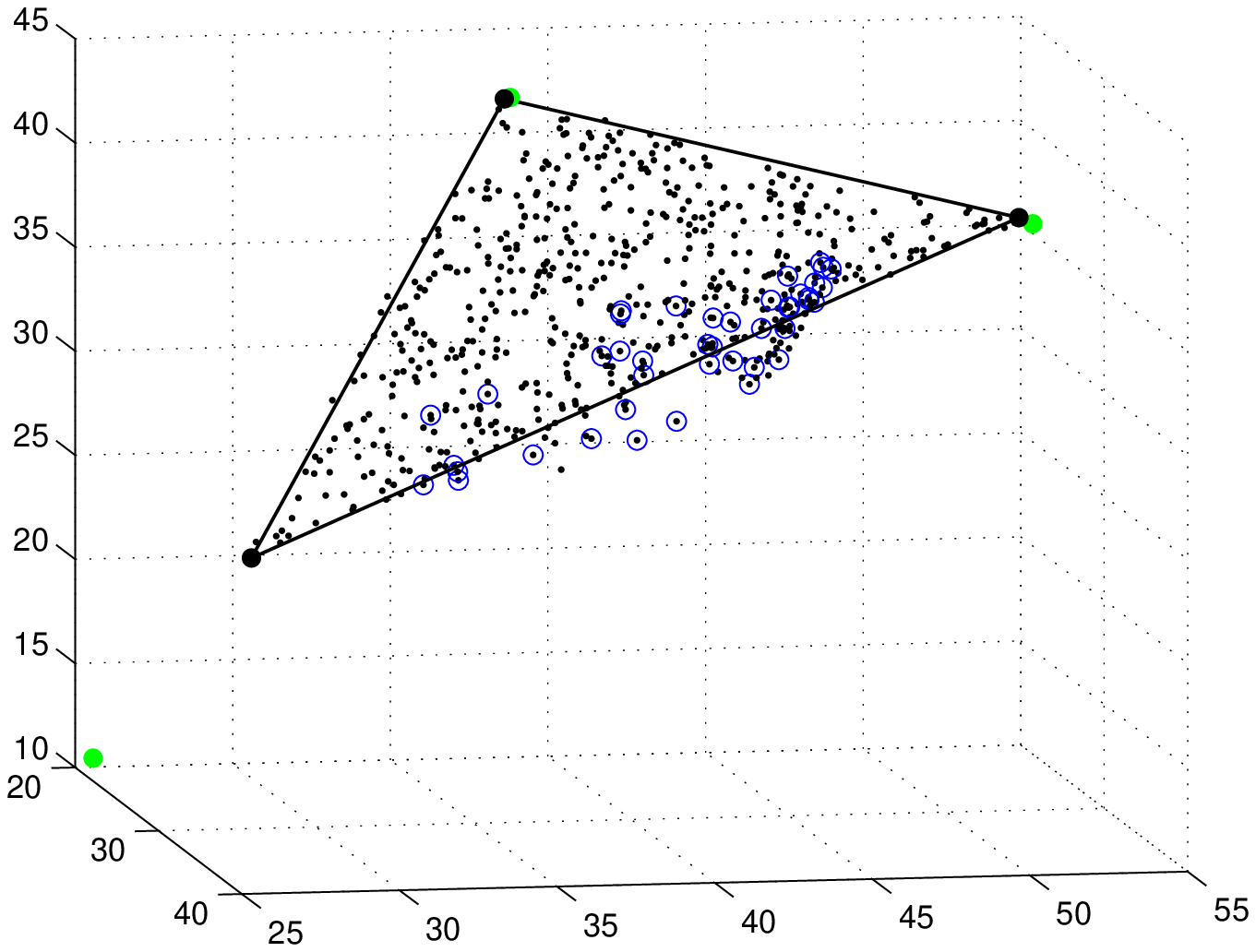}
                \caption{Seventh iteration.}
                \label{fig:itt7}
        \end{subfigure}
        ~
        \begin{subfigure}[b]{0.45\textwidth}
                \includegraphics[width=\textwidth]{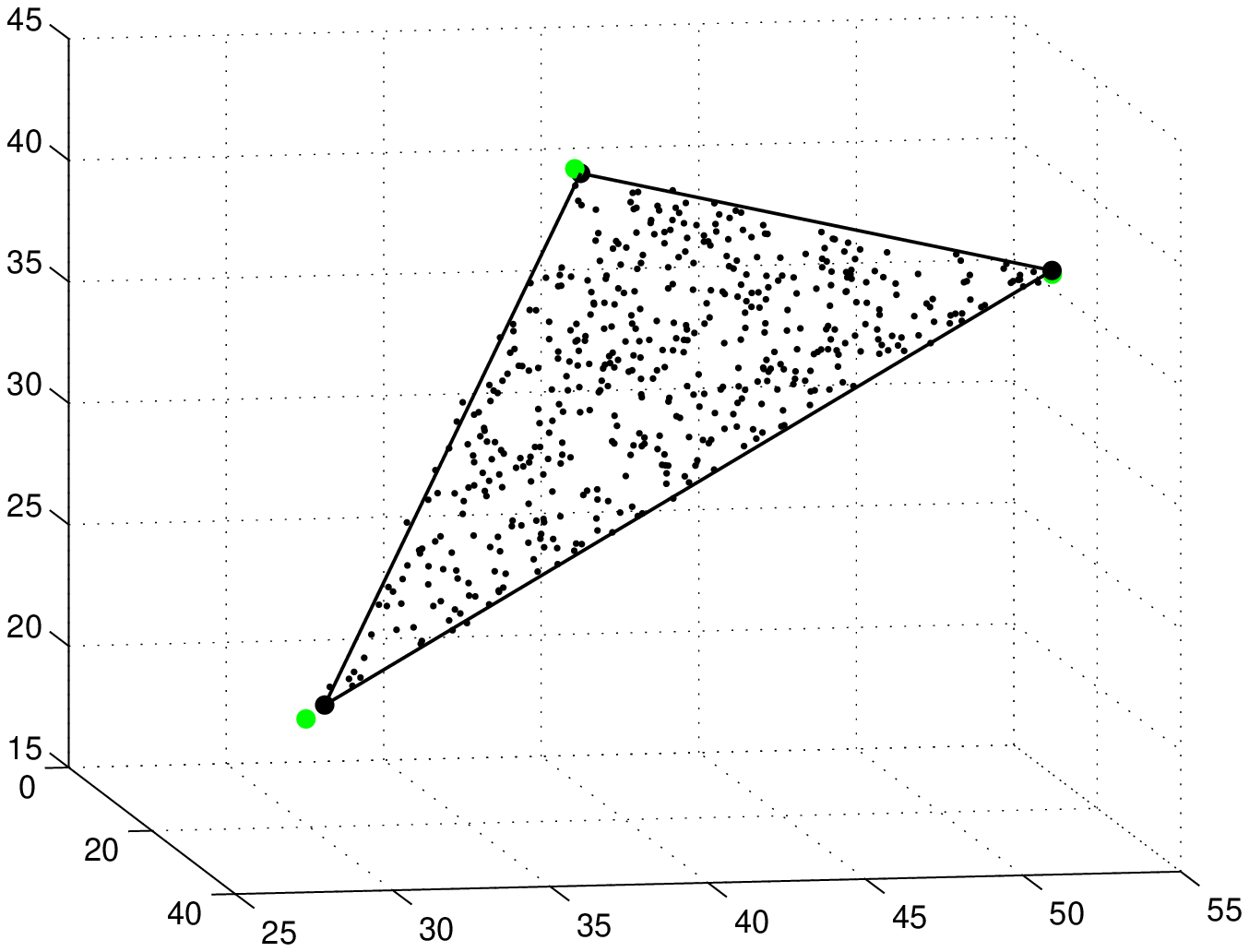}
                \caption{Final result (after 10 iterations).}
                \label{fig:itt10}
        \end{subfigure}        
        \caption{Graphical illustration of the endmember estimation process using the proposed iterative algorithm. The data set consists of 2000 pixels, with a proportion of 50\% nonlinearly mixed pixels obtained with the GMB model and $\eta_d = 0.5$. Green dots are the current estimated endmembers, and black dots are the data projected onto the subspace spanned by the columns of the current matrix $\MM$. The true endmembers are shown as black circles at the vertices of the true simplex drawn with black lines. The data discarded at the corresponding iteration are shown within blue circles.}\label{fig:Mest}
\end{figure*}

Figure~\ref{fig:Mest} illustrates a representative example of evolution obtained with the proposed iterative endmember extraction algorithm. These plots correspond to a simulation performed using 1000 synthetic samples, 500 being linearly mixed and 500 being nonlinearly mixed. The nonlinearly mixed pixels were created using the GBM~\eqref{eq:MGBM} with $\eta_d =0.5$. The data were projected onto the space spanned by the columns of the current endmember matrix $\MM$. They are represented as black dots. The current endmembers are shown as green dots. The true endmembers are shown as black circles at the vertices of the true simplex drawn with black lines. The data discarded at each iteration are shown within blue circles. Figure~\ref{fig:Mest}a shows the first iteration of Algorithm~\ref{alg:MVIEE}. Numerous nonlinear samples are outside the simplex and endmember are poorly estimated. The situation improves in Fig.~\ref{fig:Mest}b, which depicts the fourth iteration.  Here, much less data lie outside the simplex, and two of the endmember estimates have improved significantly. Similar improvement can be noticed in the seventh iteration in Fig.~\ref{fig:Mest}c. The final result obtained after 10 iterations only is shown in Fig.~\ref{fig:Mest}d, where most of the nonlinear data were discarded and the endmember estimates are clearly close to the true endmembers.

\subsection{Real Data}

\begin{figure*}
 \centering
 \begin{subfigure}[b]{0.45\textwidth}
 \includegraphics[width=\textwidth]{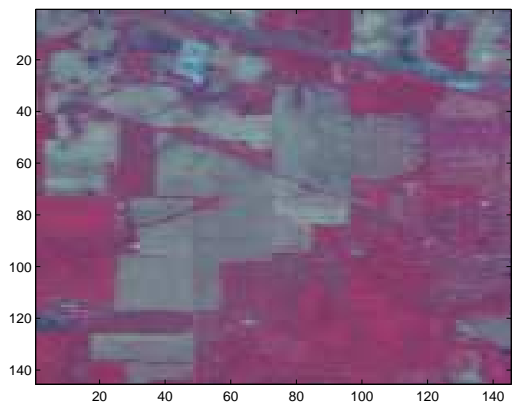}
 \caption{Indian Pines representation (3-band combination \#1)}\label{fig:indianPinesRep}
 \end{subfigure}%
 ~
 \begin{subfigure}[b]{0.45\textwidth}
  \includegraphics[width=\textwidth]{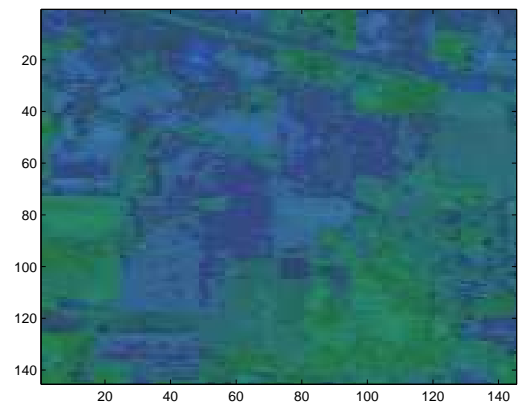}
  \caption{Indian Pines representation (3-band combination \#2)}\label{fig:indianPinesRep2}
 \end{subfigure}
 \caption{Indian Pines test site representation selecting 3 different bands in (a), and 3 other bands in (b). }
\end{figure*}

\begin{figure*}
 \centering
 \begin{subfigure}[b]{0.4\textwidth}
   \includegraphics[width=\textwidth]{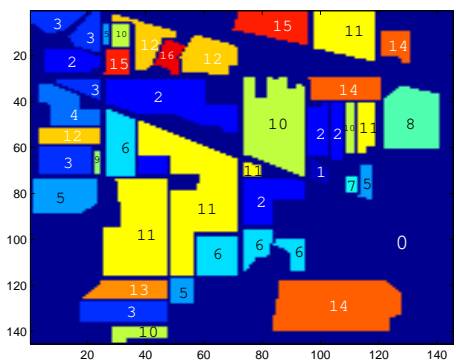}
   \caption{Indian Pines ground truth.}\label{fig:indianPinesGT}
 \end{subfigure}
~
\begin{subfigure}[b]{0.4\textwidth}
   \includegraphics[width=\textwidth]{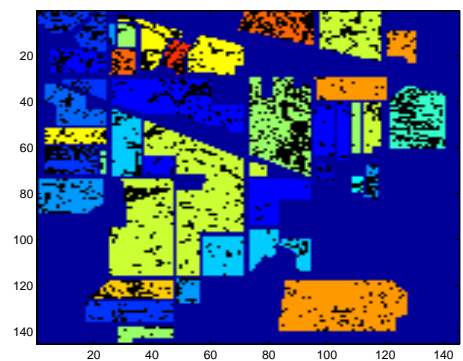}
   \caption{Indian Pines detection map.}\label{fig:indianPinesDetMap}
 \end{subfigure}
 \caption{Detection of nonlinearly mixed pixels in Indian Pines hyperspectral image. Black pixels were detected as nonlinearly mixed ones by the proposed detector.}
\end{figure*}

To test the proposed detector using real images, we used the data set available at the Indian Pines test site in North-western Indiana~\cite{Dopido-2011-ID316}. This image was captured by the AVIRIS (Airborne Visible/Infrared Imaging Spectrometer). It has $145\times 145$ samples over 220 contiguous bands with wavelengths ranging from 366 to 2497 nm. Prior to analysis, noisy and water absorption bands were removed resulting in a total of 200 bands that were uniformly decimated to 50 to speed up simulations. The data set has a ground truth map that divides the samples into 16 mutually exclusive classes. In Table~\ref{tab:IPClasses}, the classes are organized by numbers (1 to 16), and the number of samples of each class is shown. Note, however, that the number of samples in each class can vary considerably. Note also that some classes are composed of different materials. We can count 20 different materials if we consider grass as an isolated material for the whole image. We chose to count each grass (depending on the accompanying material) as a different material,  leading to 22 endmembers. Figures~\ref{fig:indianPinesRep} and~\ref{fig:indianPinesRep2} display images from the Indian Pines region constructed by selecting three different bands, while Fig.~\ref{fig:indianPinesGT} presents the ground truth map for this image, where each class is represented by a different color. In Figure~\ref{fig:indianPinesGT}, we also indice the class number for each area, where 0 represents the background, which is an unclassified area.

To perform the simulations, we divided the image into eight sub-images to work with smaller areas of the image and to deal with 3 to 4 endmembers at a time. To define these sub-images, we also paid attention to balance the number of samples per endmember. By looking at Figs~\ref{fig:indianPinesRep} and~\ref{fig:indianPinesRep2}, we can note that some classes seem to have materials that are not accounted for in the available ground-truth information. For instance, this is the case for classes 5, 11 and 14. Therefore, we introduced extra endmembers for some of the sub-images. Table~\ref{tab:subimage} describes how the sub-images were organized, showing the classes, materials, numbers of pixels and endmembers chosen for each of the eight sub-images. 

For each sub-image, we estimated the endmembers as discussed in Section~\ref{sec:EMExtraction}, with $N_{\max}=10$, a relaxing factor initially set to $r_f = 0.8$, and incremented by $r_{\text{inc}}=(1-r_f)/N_{\max}=0.2$ at each of the 10 iterations. Then, we ran the detection algorithm with $\text{PFA} = 0.001$. We performed the unmixing step using FCLS for pixels detected as linearly mixed and  SK-Hype for pixels detected as nonlinear mixtures. Figure~\ref{fig:indianPinesDetMap} presents the detection map superimposed to the ground-truth classes, where black dots represent pixels detected as nonlinearly mixed. 

Comparing the detection map in Fig.~\ref{fig:indianPinesDetMap} with Figs~\ref{fig:indianPinesRep} and~\ref{fig:indianPinesRep2}, one can note similarities between the detection map and some patterns observed in the image representations. For instance, the triangular shape in class 11 in Fig.~\ref{fig:indianPinesDetMap} is just besides what seems to be a road or trail when looking to Figure~\ref{fig:indianPinesRep}. Similarities can be found between contours of detected nonlinear regions in Fig.~\ref{fig:indianPinesDetMap} and the corresponding regions in Figs~\ref{fig:indianPinesRep} or~\ref{fig:indianPinesRep2}. Table~\ref{tab:indianPinesReconstructionRMSE} reports the RMSEs for the reconstruction error for each of the eight sub-images using three approaches, namely FCLS, SK-Hype, and Detect-then-Unmix. The results marked in blue correspond to the lowest RMSEs. For almost all sub-images, we note that the use of a nonlinear mixture detector improved the image reconstruction when compared to the pure linear or pure nonlinear unmixing strategies.

\begin{table}
\caption{Indian Pines classes by region.}\label{tab:IPClasses}
\begin{center}
\begin{tabular}{ccc}
Class number & Class & Num. of Samples\\
\hline
1 & Alfalfa & 46\\
2 & Corn-notill	 & 1428\\
3 & Corn-mintill & 830\\
4 & Corn & 237\\
5 & Grass-pasture & 483\\
6 & Grass-trees	 & 730\\
7 & Grass-pasture-mowed & 28\\
8 & Hay-windrowed & 478\\
9 & Oats & 20\\
10 & Soybean-notill & 972\\
11 & Soybean-mintill & 2455\\
12 & Soybean-clean & 593\\
13 & Wheat & 205\\
14 & Woods & 1265\\
15 & Buildings-Grass-Trees-Drives & 386\\
16 & Stone-Steel-Towers & 93\\\hline
\end{tabular}
\end{center}
\end{table}

\begin{table*}
\caption{Subimages organization}\label{tab:subimage}
\centering
{\renewcommand{\arraystretch}{1.2}
 \begin{tabular}{|c|c|c|c|c|}
 \hline
 Subimage & Classes & Materials & \# of pixels & \# of endmem.\\ \hline
 1 & 9 and 7 & Oats and grass-pasture-mowed & 48 & 3\\ \hline
 2 & 1, 4 and 13 & Alfafa, wheat and corn & 488 & 3\\ \hline
 3 & 16 & Stone-steel-towers & 93 & 3\\ \hline
 4 & 15 & Buildings-grass-trees-drives & 386& 4\\ \hline
 5 & 5 & Grass-Pasture & 483 & 3\\ \hline
 6 & 8 and 12 & Hay-windrowed and Soybean-clean & 1071 & 3\\ \hline
 7 & 3,6 and 10 & Corn-mintill, grass-trees and soybean-notill & 2532 & 4\\ \hline
 8 & 14 2 11 & Woods, corn-notill, soybean-mintill & 5148 & 4\\ \hline
 \end{tabular}}
\end{table*}

\begin{table*}
\caption{Indian Pines recontruction error (RMSE) by subimage.}\label{tab:indianPinesReconstructionRMSE}
\centering 
{\renewcommand{\arraystretch}{1.2}
\begin{tabular}{|c|c|c|c|}
\hline
\multirow{2}{*}{Subimage} & \multicolumn{3}{c|}{RMSE $\pm$ STD}\\
\cline{2-4}
& FCLS & SK-Hype & Detect-then-Unmix\\
\hline
1 & \cblue{0.0028627 $\pm$ 6.6939e-06} & 0.0030332 $\pm$ 6.0053e-06 & 0.0029083 $\pm$ 6.5229e-06\\
2 & 0.0038963 $\pm$ 1.2293e-05 & 0.003881 $\pm$ 9.4813e-06 & \cblue{0.0038391 $\pm$ 1.1505e-05}\\
3 & 0.0044259 $\pm$ 2.9087e-05 & 0.0035981 $\pm$ 8.9722e-06 & \cblue{0.0035537 $\pm$ 9.8622e-06}\\
4 & 0.0040145 $\pm$ 1.1417e-05 & 0.0039097 $\pm$ 8.0165e-06 & \cblue{0.0038895 $\pm$ 8.5058e-06}\\
5 & 0.0030848 $\pm$ 7.0516e-06 & 0.0032353 $\pm$ 5.9761e-06 & \cblue{0.0030527 $\pm$ 6.2275e-06}\\
6 & 0.0039905 $\pm$ 6.5627e-06 & 0.004055 $\pm$ 7.1531e-06 & \cblue{0.0039644 $\pm$ 6.6603e-06}\\
7 & 0.0034804 $\pm$ 5.8657e-06 & 0.0035049 $\pm$ 5.9207e-06 & \cblue{0.0034552 $\pm$ 5.9632e-06}\\
8 & 0.0037665 $\pm$ 7.5723e-06 & 0.0039314 $\pm$ 7.3092e-06 & \cblue{0.0037531 $\pm$ 7.4932e-06}\\
\hline
\end{tabular}}
\end{table*}

\section{Conclusions}\label{sec:conc}

This paper proposed a nonparametric method for detecting nonlinear mixtures in hyperspectral images. The performance of the detector was studied for supervised and unsupervised unmixing problems. Additionally, an iterative algorithm was derived for endmember estimation as a pre-processing step for unsupervised unmixing problems.  It was shown that the combined use of the proposed detector and endmember estimation algorithm leads to better unmixing results when compared to state-of-the-art solutions. A degree of mixture nonlinearity based on the relative energies of the linear and nonlinear contributions to the mixing process was defined to quantify the importance of the linear and nonlinear model counterparts. Such a definition is important for a proper evaluation of the relative performances of different nonlinear mixture detection strategies.

  \clearpage

\bibliographystyle{IEEEbib}
\bibliography{hyperspectral}



\end{document}